\documentclass[lettersize,journal]{IEEEtran}
\usepackage{amsmath,amsfonts}
\usepackage{algorithm}
\usepackage{array}
\usepackage{textcomp}
\usepackage{stfloats}
\usepackage{url}
\usepackage{verbatim}
\usepackage{graphicx}
\usepackage{cite}

\usepackage{hyperref}
\usepackage{url}
\usepackage[numbers]{natbib}
\usepackage{algorithm}
\usepackage{algpseudocode}

\usepackage{multirow}
\usepackage{graphicx}

\usepackage{caption}
\usepackage{subcaption}
\usepackage{booktabs,siunitx,tabularx,subcaption}

\begin{document}

\title{A Large-scale Analysis on the Use of Arrival Time Prediction for Automated Shuttle Services in the Real World}

\author{Carolin~Schmidt,
        Mathias~Tygesen, and Filipe~Rodrigues
\thanks{The authors are with the Department of Technology, Management
and Economics, Technical University of Denmark (e-mail: csasc@dtu.dk;
mnity@dtu.dk; rodr@dtu.dk). This work was supported by Horizon 2020 Research and Innovation Program under Grant 875530. \\ © 2025 IEEE. Personal use of this material is permitted. Permission from IEEE must be obtained for all other uses, in any current or future media, including reprinting/republishing this material for advertising or promotional purposes, creating new collective works, for resale or redistribution to servers or lists, or reuse of any copyrighted component of this work in other works. Digital Object Identifier 10.1109/TITS.2025.3643319}
}
\maketitle

\begin{abstract}
Urban mobility is on the cusp of transformation with the emergence of shared, connected, and cooperative automated vehicles. Yet, for them to be accepted by customers, trust in their punctuality is vital. Many pilot initiatives operate without a fixed schedule, enhancing the importance of reliable arrival time (AT) predictions. This study presents an AT prediction system for automated shuttles, utilizing separate models for dwell and running time predictions, validated on real-world data from six cities. Alongside established methods such as XGBoost, we explore the benefits of leveraging spatial correlations using graph neural networks (GNN). To accurately handle the case of a shuttle bypassing a stop, we propose a hierarchical model combining a random forest classifier and a GNN. The results for the final AT prediction are promising, showing low errors even when predicting several stops ahead. Yet, no single model emerges as universally superior, and we provide insights into the characteristics of pilot sites that influence the model selection process and prediction performance. Finally, we identify dwell time prediction as the key determinant in overall AT prediction accuracy when automated shuttles are deployed in low-traffic areas or under regulatory speed limits. Our meta-analysis across six pilot sites in different cities provides insights into the current state of autonomous public transport prediction models and paves the way for more data-informed decision-making as the field advances.
\end{abstract}


\section{Introduction}
%
%
%
%


\IEEEPARstart{S}{hared}, connected, and cooperative autonomous vehicles offer a unique opportunity for a fundamental change in urban mobility. They can provide seamless door-to-door mobility of people and freight delivery services, which can lead to more accessible, greener, and more sustainable cities - provided they are integrated into an effective public transport system. 
For users to accept this new mode of transportation, they need to trust its punctuality, making reliable AT prediction systems crucial for a satisfactory user experience. Fortunately, connected autonomous vehicles also constitute a unique Big Data source. When coupled with the latest developments in Machine Learning and Artificial Intelligence, this data can enable the implementation of AT prediction systems. However, an important question arises: do the findings about established machine learning models for conventional buses transfer to their automated counterparts? 
With the recent developments in both technological and regulatory matters, autonomous public transport is still in pilot testing and is constrained by local regulations. The data that these pilot projects yield, though invaluable, is limited and varies in quality. Thus, it is unclear if this data can serve as a foundation for reliable prediction models. Moreover, so far, the primary function of automated shuttles in real-life case studies has been to connect existing public transport trunk lines with remote areas, such as schools, residential care facilities, and shopping centers. Typically, there is no historical data, such as passenger counts, from existing infrastructure that can be used as an indication for this new service. Furthermore, these shuttles often operate without fixed schedules, rendering schedule adherence infeasible and thus making AT prediction especially challenging, yet, at the same time, an even more important task.

To our knowledge, this paper is the first to systematically benchmark a broad spectrum of AT prediction models for automated shuttles, while explicitly separating and modeling running-time and dwell-time components. Because automated, smaller shuttle services are often deployed to serve previously remote areas, demand is highly volatile, and thus dwell-time data are heavily zero-inflated: stops are often skipped when no passengers are present. To handle this, we introduce a hierarchical framework of a Random-Forest classifier and a Graph Neural Network regressor. Further, this work is, to our knowledge, the first use of GNNs for dwell-time prediction, exploiting spatial correlations between stops, whereas earlier studies rely on simpler models \cite{Rashidi2013ApproximationAS, Chen2007UsingAP, MA2019536, PETERSEN}. We evaluate all models on real-world data from six heterogeneous pilot sites, spanning different vehicle types, routes, and operating contexts, thereby providing insights that are valuable to both researchers and practitioners working with emerging automated-shuttle systems.


First, we review related work on bus AT prediction. Next, we present our segment-based approach for separating dwell and running time, along with the corresponding prediction models. We then describe the data pre-processing steps before presenting the final results. Finally, we conclude and outline future research directions.


\section{Related work}
Bus AT prediction is a large research area within Intelligent Transport Systems. Historically, Kalman Filters have been a common approach to AT predictions due to their ability to maintain states between predictions and filter out noise \cite{Chen2004ADB, Wall1998AnAF, Sinn2012PredictingAT, Achar2020Kalman}. More recently, Machine Learning (ML) models have demonstrated promising performance for AT prediction tasks. Shallow ML techniques such as Support Vector machines \cite{Yu2006BusAT} and Random Forests \cite{Li2017BusAT} have been explored; however, most work has focused on models based on Artificial Neural Networks (ANN). For instance, \citep{Lin2013RealTimeBA} developed a hierarchical ANN model, which significantly outperformed Kalman Filters for short-distance routes. Kalman filters remain attractive for their recursive updates and explicit uncertainty measurement, but recent studies show that deep ML models (LSTM, XGBoost) and, especially, GNNs offer benefits on bus AT prediction by exploiting high-dimensional context and network-wide spatial correlations, with hybrid ML + KF schemes now emerging as a promising combination \cite{Bai, Achar,Shen} Long Short-Term Memory models (LSTMs) have become popular for AT prediction due to their ability to capture the non-linear behavior of time-series data, effectively modeling temporal dependencies \cite{Agafonov2019BusAT, Pang2019LearningTP, Lingqiu2019ALB}. In \cite{PARSLOV2023}, the authors integrate uncertainty estimation into recurrent neural networks to predict running times, improving decision-making for transfer synchronization. In \cite{PETERSEN}, the authors combine LSTMs with Convolutional Neural Networks to capture both the temporal and spatial correlations between bus stops. \citep{Ma2022MultiattentionGN} uses Graph Neural Networks (GNNs) for city-wide bus travel time estimation. However, unlike our approach, their model does not separate travel time into dwell and running time. 

In \cite{Chen2007UsingAP}, the authors, using simple ANNs, identify dwell time as among the most important factors determining AT using simple ANNs. As such, it can be beneficial to split the task into predicting the dwell and running time separately. In \cite{MA2019536}, the authors propose a segment-based approach to predict bus AT by dividing bus routes into dwelling and transit segments. The final AT predictions obtained by shallow ML approaches or simple ANNs are then generated by accumulating the independent predictions. Similarly, \citep{PETERSEN} advocate for splitting bus AT models into travel and dwell time models, utilizing LSTM and CNNs for the running time predictions and a simpler exponential smoothing model for dwell time predictions. For a more in-depth survey of general AT prediction research, we refer to \cite{Singh2022ARO}.

As automated shuttles are a new technology with limited real-world deployment, the research on predicting AT for automated shuttles is lacking. \citep{Antypas} did an initial study of the running time of automated shuttles. They used factors such as location, elapsed time, speed, and acceleration as inputs for a Gradient Boosting Regression Tree model to predict the remaining segment time. The authors added artificial stops to the route as the study was done in the early stages of an automated shuttle pilot. 

Building on data from advanced pilot sites, our work, to the best of the authors' knowledge, is the first to benchmark a range of prediction models on separate dwell and running time prediction tasks with data from automated shuttles in real-world deployments.
Moreover, our approach is the first to use GNNs and hierarchical GNN architectures to separately predict dwell and running times, incorporating spatial information into dwell time prediction - a component overlooked in previous research. 

\section{Methods}

\begin{figure*}[!t]
    \centering
    \includegraphics[width=0.65\linewidth]{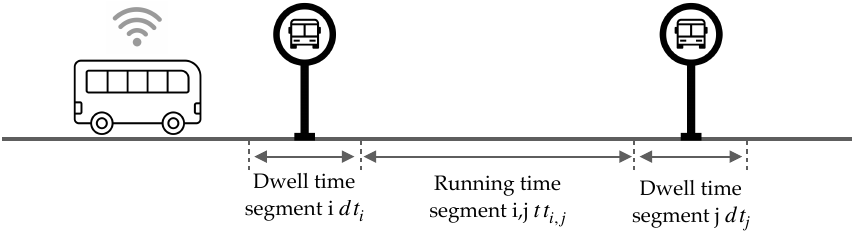}
    \caption{Our segment-based approach. The route is divided into dwell and running time segments.}
    \label{fig:segment_based}
\end{figure*}

Dwell times are mainly impacted by the uncertain number of boarding passengers and hence show structural differences to running times, which are primarily impacted by traffic conditions. Therefore, in this work, we adopt a segment-based approach, employing two separate models to predict the running and dwell times. Figure \ref{fig:segment_based} provides an overview of the segment-based approach. The dwell time corresponds to the time the shuttle is stationary at a shuttle stop, and the running time to the time it travels between two stops, including acceleration and deceleration time.
To obtain the final travel time $T_{i,j}$ of the automated shuttle from stop $i$ to stop $j$, we aggregate the predictions for the individual segments for the dwell time $dt$ and running time $tt$ along the route: 
    \begin{equation}
        T_{i,j} = \sum_{k=i}^{j-1} tt_{k, k+1} + \sum_{k=i+1}^{j-1} dt_k. \nonumber
    \end{equation}

\begin{figure*}[!t]
    \centering
    \includegraphics[width=0.8\linewidth]{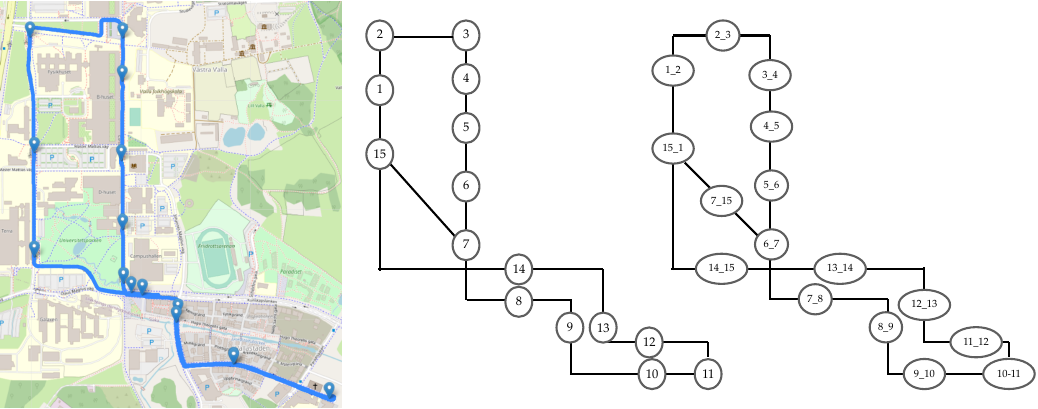}
    \caption{Construction of the graph for GNN predictions from the original route (left) to the final graph for dwell time (middle) and running time predictions  (right)}
    \label{fig:Graph_lin}
\end{figure*}

\subsection{Dwell and running time prediction models}
\label{sec:pred_models}
Given that current data on automated shuttles is confined to test pilots, our models must operate on limited input. Many studies have equipped their models with real-time traffic information from available car data, e.g., taxi data \cite{MA2019536}, to obtain accurate running time predictions. However, as automated shuttles often operate on both shared public roads and private lanes or share lanes only with pedestrians and cyclists, leveraging standard road traffic speed estimates becomes unfeasible. Moreover, these estimates might not be as impactful in the case of automated shuttles, as they adhere to specific speed regulations designed for such vehicles, which are often lower than the road's general speed limit. Hence, our approach adopts a time-series-based approach, where we include previous observations of the running and dwell time, respectively, on the segment into the regression task. However, given the data sparsity resulting from pilot testing, the most recent observations can quickly become outdated and, thus, less reliable. Hence, our models only include the two previous observations per segment. As such, our features include the current position of the shuttle, current weather data, current time, and the ID of the segment and the ID of the vehicle in question (since vehicles can have different manufacturers and therefore different behaviors), along with the two most recent observations. Given vehicle id $i$, segment id $j$ and time $t$, the input to our prediction models is:
\begin{equation}
    \boldsymbol{x}_{i,j,t} = \left[\tau_t, w_t, v_i, s_j, p_j, l^{(1)}_{j,t}, l^{(2)}_{j,t} \right],\nonumber
\end{equation}
where $\tau_t$ are the geometrically transformed time-of-day and day-of-week features, $w_t$ is a vector representing the temperature, precipitation, and wind speed at an hourly resolution, $v_i$ is a one-hot encoding of the vehicle id, $s_j$ is a one-hot encoding of the segment id, $p_j$ the position of the stop or start of running time segment (latitude and longitude), and $l^{(1)}_{j,t}$ and $l^{(2)}_{j,t}$ are the most recent and second most recent lag of segment $j$ at time $t$. The lags are either historical dwell time or running time, depending on which is being predicted. Then, we define a model $\mathcal{F}$ that takes in the input data $x_{i,j,t}$ and predicts the next running or dwell time for that vehicle on the segment, 
\begin{equation}
    \boldsymbol{x}_{i,j,t} \xrightarrow{\mathcal{F}} \hat{y}_{i,j,t}.\nonumber
\end{equation}
Specifically, $\hat{y}_{i,j,t}$ corresponds to the predicted running time on segment $j$ given entry time at $t$ or dwelling time at stop $j$ given arrival at the stop $j$ at time $t$. 
In graph neural network-based approaches, where the input must be structured as a graph with nodes and their respective features, the input becomes a matrix $\boldsymbol{X}$ where the $k$'th row is $x_{i,k,t}$. The GCN model then predicts a vector, $\boldsymbol{y}_{i,t}$, with a prediction for each node. The prediction for segment $j$ can then be taken out as the $j$'th position.

Specifically, for the GNN, the input data must be structured as a graph with nodes and edges linking these nodes based on an adjacency matrix. For the dwell time model, each node represents a stop. For the running-time model, each node represents the road segment between two consecutive stops. The connectivity of the input graph corresponds to the physical shuttle routes. Two nodes are connected if, in at least one of the shuttle routes, they are visited consecutively. We illustrate the graphs for Linköping in Figure \ref{fig:Graph_lin}. 

Our feature set is designed to allow temporal generalization, using temporal and weather information. This approach addresses the most common scenario that pilot sites aim to evaluate. For cross-city transferability, we anticipate that either initial fine-tuning on a dataset from the new city or an extension of the current feature set, with e.g., demand estimates, POI densities, or road types, to be necessary.

To evaluate the different prediction models, we compare their performance against two naive baseline models:
\begin{itemize}
    \item a \textit{Lag model}, which predicts the most recent observation on the respective segment;
    \item a \textit{Mean regressor}, also often referred to as a historical average model, which predicts the mean of the respective segment.
\end{itemize}
We employ machine learning techniques that have proven successful in predicting both the dwell and the running times of conventional buses. Studies highlighted the competitive performance of tree-based methods, especially XGBoost, in both dwell time \cite{10125868, Rashidi_dwell}, travel time \cite{travel_time_xgboost}, and running time predictions \cite{travel_time_xgboost2}. We include the following tree-based models:
\begin{itemize}
    \item \textit{XGBoost} \cite{xgboost}, which stands for eXtreme Gradient Boosting, is an optimized gradient boosting algorithm that constructs decision trees sequentially, with each new tree aiming at minimizing the residual errors of the previous ones. 
     \item \textit{Random Forest} \cite{RF}, a simpler yet highly effective tree-based model. Random Forest is an ensemble method that builds multiple decision trees and aggregates their results to improve prediction accuracy and robustness.
\end{itemize}

\begin{table*}
        \small
        \addtolength{\tabcolsep}{-2.5pt}
	\caption{Pilot site characteristics. }\label{tab:show1}
	\begin{center}
		\begin{tabular}{l c c c c c c c}
	Site   & Country & \#Routes & \#Stops     & \#Vehicles  &Shuttle types (\#types, \#passengers) & Speed limit &Deployment  \\\hline
        Linköping  & Sweden  &2         &15           &3            & Minibus (2, 11/13)&30-40km/h&Campus+Residental\\
        Tampere    & Finland &1         &\phantom{0}7 & 2           &Van (1, 7)&40km/h&Sub-urban+Residential\\
        Les Mureaux& France  &3         &\phantom{0}7 &2            &Minibus (1, 12)&30-50km/h&Industrial area\\
        Madrid     & Spain   &1         &\phantom{0}5 &2            & Minibus (2, 30/105)&NA&Bus depot\\
        Graz       & Austria  &2        &\phantom{0}4 &1            &Passenger car&30km/h&Retail area\\
        Carinthia& Austria  &  2       &\phantom{0}8 &1             &Minibus (1, 11)&30-50km/h&Residental areal\\
			\hline
		\end{tabular}
	\end{center}
\end{table*}
\newpage
In addition, we examine three deep learning models: 
\begin{itemize}
    \item A \textit{Multi-Layer Perceptron} (MLP) that consists of fully connected layers, capturing non-linear relationships in the data through layer-wise transformations. It provides a baseline deep-learning approach for non-graph inputs and has been traditionally applied to both tasks \cite{Rashidi_dwell, GURMU201445}. 
    \item A \textit{Graph Convolutional Network} (GCN) \cite{kipf2017semi}, which propagates information among nodes in a graph using a permutation-invariant propagation rule. This enables GCNs to capture spatial dependencies within the data, making them well-suited for tasks with spatially correlated features. As a result, GCNs have proven successful in numerous transportation applications, such as traffic flow prediction \cite{Li2017DiffusionCR}, and fit our prediction tasks as both inherently include spatial correlations.  
    \item A hierarchical model that combines a Random Forest classifier with a GCN regressor (\textit{RF-GCN}). In smaller shuttle services, demand volatility yields a dataset with a potentially high number of zero dwell times. Accurately detecting these zeros is essential for AT prediction, as overestimating the dwell time leads to an overestimation of the total travel time. To address this, we propose the RF-GCN model. Drawing from the tools of zero-inflated regression, the binary RF classifier is used to detect zero observations, while the downstream GCN regresses the non-zero observations. Concretely, the RF classifies whether the shuttle will stop vs. skip using binary labels ($z_i = 1, \text{if } y_i > 0$ and $z_i = 0, \text{if } y_i = 0$). Accordingly, it is trained on a modified classification dataset $D_{\text{cls}} \;=\; \{(x_i, z_i) : i=1,\dots,N\}$. The GNN regressor predicts the duration of the dwell event $\tilde y_i$ conditional on stopping. It is trained on a subset of the original dataset, only including non-zero targets $D_{\text{reg}} \;=\; \{(x_i, y_i): y_i > 0\}$. Excluding zeros focuses the regressor on actual boarding and alighting durations and avoids bias from zero-inflation. Crucially, both stages are trained independently on their respective datasets. At inference, the classifier outputs the probability of a stop being served; if the predicted class is $\hat z_i=0$ ("skip"), we set $\hat y_i=0$. If $z_i=1$ ("stop"), we query the GCN for a continuous estimate of the dwell time $\tilde y_i$ and set $\hat y_i=\tilde y_i$.
\end{itemize}

We conduct grid searches (RF, XGBoost) and random searches (MLP, GCN) on
validation sets to optimize hyperparameters. For simpler models (RF, XGBoost), we select hyperparameters for each pilot site. For the deep learning models, due to higher computational time, we use the same parameters across sites. 
The hyperparameters and training scripts for all evaluated models are available. \footnote{https://github.com/carolinssc/Arrival-time-prediction-autonomous-shuttles}

\begin{table*}
        \small
        \addtolength{\tabcolsep}{-2.5pt}
	\caption{Summary data statistics for each pilot site. \#Observation indicates running/ dwell times. Running and dwell time stats present mean values and coefficient of variation in seconds. For dwell time, the fraction of zero observations is additionally provided in parentheses.}\label{tab:show2}
	\begin{center}
		\begin{tabular}{l c c c c c c c}
		Site  & Dates      & Avg. speed  &Max speed  & \#Observations                                      &\#Passengers   & Running time stats & Dwell time stats\\\hline
        Linköping  & 12/21-02/23 &\phantom{0}7.6 km/h  &22 km/h           &45\,059/48\,679                     &17\,683         &120, 0.14 & 28.0, 0.38, (0.06)\\
        Tampere    & 12/21-03/22 &\phantom{0}8.5 km/h  &64 km/h           &\phantom{0}6\,177/\phantom{0}6\,696 &\phantom{0}8\,643 &\phantom{0}91, 0.17 & 17.4, 1.61, (0.44)\\
        Les Mureaux & 11/22-02/23 &\phantom{0}2.9 km/h &\phantom{0}6 km/h &\phantom{0}4\,580/\phantom{0}4\,847 &\phantom{0}1\,084 &174, 0.13 & 31.3, 0.41, (0.17)\\
        Madrid     & 06/22-02/23 &\phantom{0}3.0 km/h  &17 km/h           &\phantom{0}6\,166/\phantom{0}7\,096 &\phantom{0}6\,363 &\phantom{0}63, 0.34 & 31.2, 2.70, (0.75)\\
        Graz       & 10/22-06/23 &\phantom{0}4.5 km/h  &17 km/h           &\phantom{00}\,662/\phantom{00}\,359 &\phantom{00}520   &132, 0.18 & 15.3, 1.05, (0.17)\\
        Carinthia   & 06/22-12/22&\phantom{0}8.1 km/h  & 18 km/h          &\phantom{0}1\,645/\phantom{0}1\,643 & \phantom{000}NA  &146, 0.16 & 24.0, 1.00, (0.26)\\
			\hline
		\end{tabular}
	\end{center}
\end{table*}

\section{Data}
\label{sec:data}
The data were gathered from six cities conducting pilot tests with automated shuttles, all of which are part of the SHOW (SHared automation Operating models for Worldwide adoption) project \cite{SHOWweb}. The SHOW Project is a large-scale EU initiative aiming to demonstrate the viability and effectiveness of shared, connected, cooperative, and electrified automated vehicles (AVs) in real-world urban demonstrations across 20 cities in Europe. Altogether, the project deploys a fleet of more than 80 SAE L4/L5 AVs for both passenger and cargo transport in dedicated lanes as well as mixed traffic. In this paper, we test our models on data from six of these pilot sites (Tables \ref{tab:show1}), where tests have been ongoing for several months. Currently, all vehicles are automated at level 4. The pilot sites in Table \ref{tab:show1} represent not only different stages of deployment but also some have a particular focus, such as teleoperation or the challenge of interfacing with different road users. The variability in sample sizes, induced by the different pilot lengths and data collections, allows us to analyze the impact of sample size and operating conditions on the predictive performance of the different models. In particular, limited sample sizes are something that real-world pilot projects introducing new services inevitably face in practice, and the datasets used in this study align our research with real-world conditions.

The quality of the provided data varies with the pilot site; while all provide GPS positioning and speed data, some have deviations. For instance, the speed data from Linköping is noisy, and in Les Mureaux, the sampling rate of the GPS data is decreased mid-pilot. Such inconsistencies are not unique to our study but are representative of many other real-world pilots, both current and upcoming. As shown in Tables \ref{tab:show1} and \ref{tab:show2}, Linköping stands out as the site with the longest ongoing pilot, as well as the longest route, contrasting with Graz, which has the shortest route and a very limited number of observations and passengers. The majority of pilots deploy minibusses, with the exceptions of Tampere and Graz, which use vans and passenger cars, respectively. This difference is reflected in the mean dwell time, as these two sites exhibit roughly half the dwell time compared to the other pilots. While most pilots operate in suburban and residential areas, Madrid and Les Mureaux stand out. Madrid’s pilot, conducted within a private bus depot and deploying higher capacity vehicles, focuses on automated depot management and auto parking, thus yielding the highest variability in dwell and running times. On the other hand, Les Mureaux, operating for employees on a private company site, shows the second-lowest passenger count and the lowest variability in running time. These distinct pilot site characteristics provide an optimal foundation for our model evaluations since they provide a comprehensive representation of the broad array of characteristics that can be found in real-world deployments, thus allowing for an insightful meta-analysis of the results.  

\subsection{Data pre-processing}
\label{subsec:data_pre}
To enable a clear distinction between running and dwell time in public transport systems, careful data pre-processing is crucial. We provide an overview of the data preprocessing algorithm in Algorithm \ref{algo1}. 

\begin{algorithm}
\caption{Travel and Dwell Time Calculation}
\begin{algorithmic}[1]
\State \textbf{Input:} GPS data with coordinates, timestamps, and optionally speed; Shuttle routes
\State \textbf{Output:} Running times and dwell times between stops
\Procedure{Calculate Times}{data, stops}
    \State 1. Assign stop IDs to GPS points based on proximity to the next valid stops in the route.
    \State 2. Mark observation as stationary if speed $<0.01$ or GPS differences below jitter threshold.
    \State 3. For each stop, compute \textit{dwell time} as the total time the shuttle is stationary.
    \State 4. For each pair of consecutive stops, compute \textit{running time} as the time between departure and arrival.
    \EndProcedure
\end{algorithmic}
\label{algo1}
\end{algorithm}

 We consider two scenarios: One where we have speed and GPS data, and another where we rely solely on the latter. In the first scenario, with the combination of speed and GPS data, we have explicit information about when a shuttle is in motion or stationary. To determine the dwell time, we define a shuttle as dwelling when it is at a standstill (speed of 0 km/h) and within a predetermined radius of a stop. As soon as the shuttle starts moving, we start the running time calculation for the corresponding route segment. If the shuttle bypasses a stop without stopping, we consider the dwell time as zero. 

However, in some cases, e.g., in Linköping, GPS positioning might be the only data available, and the information about the shuttle's movement is less explicit. In this scenario, we establish a threshold for GPS differences to detect when a shuttle dwells at a stop while accounting for the noise (GPS jitter) in the GPS data. Dwell times are then calculated based on periods of minimal movement, as defined by the threshold. A predefined stop radius remains equally relevant here to identify when the shuttle is at a stop.

There are additional cases that require specific attention in our pre-processing. There are instances when a shuttle is parked between runs or traveling to and from its depot without turning off the GPS device. We manually identify these relevant locations and exclude instances where the shuttle deviates from its designated route from our data set. We also handle situations where two stops on opposite sides of the road have overlapping radii by hard-coding all possible stop orders in the pre-processing. This way, we can precisely determine the relevant shuttle stop for each data point, eliminating ambiguity and potential errors. Other site-dependent adjustments must be considered, such as the change in GPS sampling frequency in Les Mureaux. Here, the GPS sampling rate was reduced mid-pilot, affecting time calculations, which requires us to discard the impacted data for consistency. These aspects are not unique to our case study but represent challenges commonly encountered in real-world deployments. We emphasize the importance of addressing them to ensure the most accurate data preprocessing possible.

Finally, we enrich our data set with weather data at an hourly resolution and perform sine and cosine encodings of time-of-day and day-of-week features. For a correct model evaluation, we hold out a part (approximately 10 $\%$) of the data set depending on the date to avoid data leakage and test temporal generalisation. In Linköping, Tampere, Madrid, and Graz, our test set selects one month of data and one week in Les Mureaux and Carinthia.

The data pre-processing step is of utmost importance to make an accurate distinction between dwell time and running time. Since our data originates from real-world pilot tests, we found careful data cleaning to be essential to obtain a data set that reflects regular operation as accurately as possible.  


\begin{table}[]
	\caption{Results for Running Time Predictions Depending on Lags (Results are (RMSE/MAE))}\label{tab:per_vs_all_veh}
	\begin{center} 
        \addtolength{\tabcolsep}{-2pt}
        \footnotesize
	\begin{tabular}{c  c c c c}
            & \multicolumn{2}{c}{Lag Model}                                   & \multicolumn{2}{c}{XGBoost}  \\\
Site        & All vehicles                  & Per vehicle                     & All vehicles                   & Per vehicle      \\ \hline
Linköping  & 22.10/ 14.47                  & \textbf{20.22}/ \textbf{11.98}  & 15.57/ 10.15                   & \textbf{15.28}/  \phantom{0}\textbf{9.76}   \\
Tampere     & 16.40/ 11.53                  & 16.40/ \textbf{11.28}           & 12.41/ \phantom{0}8.94         & \textbf{12.19}/ \phantom{0}\textbf{8.58}\\
Les Mureaux       &\textbf{21.60}/ \textbf{11.23} & 21.88/ 11.42                    & 18.73/ 10.68                   & \textbf{18.47}/ \textbf{10.41}   \\
Madrid      &\textbf{29.14}/ \textbf{18.85} & 29.47/ 19.40                    & \textbf{21.09}/ \textbf{14.56} & 21.21/ 14.57\\ \hline

		\end{tabular}
	\end{center}
\end{table}

\begin{table*}
\centering
\caption{Results for Running Time Prediction}
\label{tab:travel_time}
\addtolength{\tabcolsep}{-2pt}
\small
\begin{tabular}{cccccccccc}
\multicolumn{1}{l}{}        &      & Lag   & Mean                             & LR                               & RF                               & XGBoost                                   & MLP                      & GCN              \\ \hline
\multirow{4}{*}{Linköping} & RMSE & 20.22 & 15.87                            & 15.90                            & 15.34                            & \textbf{15.28}                            & $16.33 \pm 0.07$         & $16.33 \pm 0.07$ \\
                            & MAE  & 11.98 & 10.47                            & 11.04                            & 9.90                             & 9.76                                     & $\textbf{9.50} \pm 0.03$ & $9.53 \pm 0.03$ \\ 
                            &MAPE & 9.37& 8.22& 8.84 & 7.73 & 7.61&\textbf{7.18} $\pm$ 0.04& 7.20 $\pm$ 0.04\\
                            & $R^2$  &  0.74 & 0.84                            & 0.84                            & \textbf{0.85}                            & \textbf{0.85}                                    & $0.83 \pm 0.00$           & $0.83 \pm 0.00$ \vspace{0.2cm}\\

\multirow{4}{*}{Tampere}    & RMSE & 16.40 & 12.43                            & 12.42                            & 12.34                            & \textbf{12.19}                            & $12.91 \pm 0.14$         & $12.97 \pm 0.10$ \\
                            & MAE  & 11.28 & \phantom{0}8.91                  & \phantom{0}8.87                  & \phantom{0}8.87                  & \phantom{0}\textbf{8.58}                  & $9.07 \pm 0.16$          & $9.31 \pm 0.19$  \\
                            & MAPE& 13.66 &10.95 & 10.90 & 10.84 & \textbf{10.37} &11.15 $\pm$ 0.33& 11.57 $\pm$ 0.27\\ 
                            &$R^2$ & 0.80 &0.88                           &0.88                              &0.88                                &\textbf{0.89}                                      &     $0.87 \pm 0.00$                   & $0.87 \pm 0.00$  \vspace{0.2cm}\\
                            
\multirow{4}{*}{Les Mureaux} & RMSE & 21.88 & 29.96                      & 21.42                            & \textbf{18.39}                   & 18.47                                     & $23.68 \pm 0.37$         & $25.54 \pm 0.92$ \\
                            & MAE  & 11.42 & 17.61                            & 13.63                            & 10.61                            & \textbf{10.41}                            & $13.64 \pm 0.18$         & $14.81 \pm 0.57$ \\
                            &MAPE & 6.32& 8.09& 7.4 & 5.33 & \textbf{5.29} &6.57 $\pm$ 0.12& 7.10 $\pm$ 0.34\\
                            &$R^2$ & 0.95  &0.91                           &0.95                              &\textbf{0.97}                             &0.96                                       &         $0.94 \pm 0.00$                 & $0.93 \pm 0.00$ \vspace{0.2cm}\\
                            
\multirow{4}{*}{Graz}       & RMSE & 20.25 & 15.99                            & 16.28                            & 17.03                            & \textbf{14.76}                            & $17.74 \pm 0.31$         & $18.52 \pm 0.98$ \\
                            & MAE  & 15.81 & 12.97                            & 12.95                            & 13.74                            & \textbf{11.59}                            & $13.38 \pm 0.22$         & $14.09 \pm 0.59$  \\
                            &MAPE &12.51 &10.00 & 9.87 &10.61 & \textbf{9.07} &10.06 $\pm$ 0.13& 10.64 $\pm$ 0.46\\
                            &$R^2$ & 0.42  &0.64                          &0.62                             &\textbf{0.69}                                 &\textbf{0.69}                                      &    $0.55 \pm 0.02$         & $0.51 \pm 0.05$\vspace{0.2cm}\\
                            
\multirow{4}{*}{Madrid}     & RMSE & 29.14 & 24.01                            & 22.96                            & 21.13                   & \textbf{21.09}                                     & $22.71 \pm 0.14$         & $22.73 \pm 0.07$ \\
                            & MAE  & 18.85 & 17.87                            & 16.84                            & \textbf{14.31}                   & 14.56                                     & $16.23 \pm 0.09$         & $16.47 \pm 0.11$ \\
                            &MAPE &38.48 &39.94 & 36.55 & \textbf{29.48} & 30.44 &36.57 $\pm$ 0.29& 37.03 $\pm$ 0.33\\
                            &$R^2$ & 0.08  &0.37                          &0.43                              &0.51                             &\textbf{0.52}                                           &$0.44 \pm 0.00$         &$0.48 \pm 0.00$\vspace{0.2cm}\\
                            
\multirow{4}{*}{Carinthia}  & RMSE & 31.52 & 26.53                           & 26.50                            & 27.16                             & \textbf{26.07}                           & $31.86 \pm 0.14$         & $29.68 \pm 1.06$ \\
                            & MAE  & 20.74 & 16.94                           & 17.16                            & 17.60                             & \textbf{16.78}                            & $22.02 \pm 0.09$        & $19.84 \pm 0.59$ \\ 
                            &MAPE &12.50 & \textbf{9.94} & 10.30 & 10.31 & 9.96 &13.92 $\pm$ 0.76& 13.67 $\pm$ 2.06 \\
                            &$R^2$ & 0.81  &\textbf{0.87}                           &\textbf{0.87}                              &0.86                               &\textbf{0.87}                                      &      $0.81\pm 0.02$        &$0.83\pm 0.00$\\
                            \hline
\end{tabular}%
\end{table*}

\section{Results}
In this section, we perform a series of experiments on the data sets presented in the previous section. First, we investigate whether it is beneficial to create historical lags based on all vehicles or per vehicle (since different vehicles may correspond to different models and manufacturers). Then we compare the performance of the different prediction models in running and dwell time prediction, and aggregate the predictions to calculate the final arrival time along observed journeys from the test data to evaluate compounding error and model bias. Lastly, we quantify the impact of deviations from the scheduled, automated service and provide some key insights for improved data collection during pilot deployments. 
As performance metrics, we report the Root Mean Square Error (RMSE), the Mean Absolute Error (MAE), the Mean Absolute Percentage Error  (MAPE), as well as $R^2$:

\begin{align}
            \text{RMSE}(y, \hat{y}) &= \sqrt{\frac{\sum_{i=0}^{N - 1} (y_i - \hat{y}_i)^2}{N}},\\
            \text{MAE}(y, \hat{y}) &= \frac{ \sum_{i=0}^{N - 1} |y_i - \hat{y}_i| }{N},\\
            \mathrm{MAPE}(y, \hat{y})
            &= \frac{100\%}{N} \sum_{i=0}^{N - 1} \left|\frac{y_i - \hat{y}_i}{y_i}\right|.\\
            \text{$R^2$}(y, \hat{y}) &= 1 - \frac{\sum_{i=0}^{N - 1} (y_i - \hat{y}_i)^2}{\sum_{i=0}^{N - 1} (y_i - \bar{y})^2}, 
\end{align}
where $y$ is the observed target values and $\hat{y}$ is the predictions for the test set with $N$ observations. 
\subsection{Choice of lag feature}
Prior to assessing the final prediction results, we need to determine whether lags are taken from the most recent observation of the individual vehicle or the whole fleet. Hence, in Table \ref{tab:per_vs_all_veh}, we compare the predictive performance of the Lag Model and XGBoost on running time depending on the lags. Pilot testing occasionally results in sparse data, making recent lags from the whole fleet potentially advantageous. If, on the other hand, the fleet is a mix of shuttles from different manufacturers, observations from different vehicles may not be as informative. As seen in Table \ref{tab:per_vs_all_veh}, the results are not consistent across sites. Les Mureaux and Tampere deploy the same type of vehicles, while the remaining sites test different manufacturers. As a consequence, we can see a clear improvement in Linköping, with the highest number of vehicles deployed, when providing lags from the respective vehicle. For the other pilot sites, the difference is less striking. In Les Mureaux, the vehicles are identical, which leads to a better lag model for the lags based on the whole fleet. Yet, XGBoost performs better at the individual vehicle level, indicating subtle differences in the data of the vehicles. Based on this analysis, we create the lags per vehicle in Linköping, Tampere, and Les Mureaux and for the whole fleet in Madrid.

\begin{figure}
    \centering
     \includegraphics[width=\linewidth]{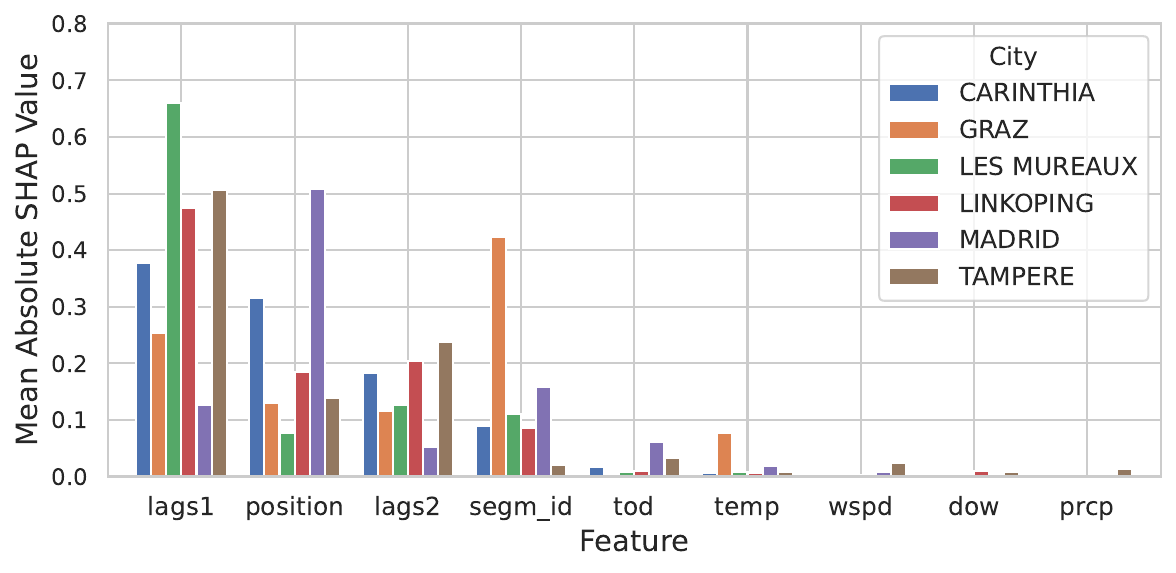}
    \caption{Mean SHAP values for XGB running time prediction. Variables include: segm\_id (segment ID), tod (time-of-day), temp (temperature), wspd (windspeed), dow (day-of-week), prcp (precipitation)} 
    \label{fig:shap_travel}
\end{figure}

\subsection{Running time prediction}
\label{sec:run_pred}
In evaluating prediction results on single segments, we first analyze the comparative performance of machine learning and deep learning models across pilot sites. In Table \ref{tab:travel_time}, we present the performance for running time predictions across models, excluding the RF-GCN model, as zero observations are impossible. For most pilot sites, the tree-based methods RF and XGBoost are superior, with XGBoost outperforming in Tampere and Graz, RF in Madrid, and the performance being equal in Les Mureaux. Only for Linköping, the ANN models outperform in MAE, with both the MLP and the GCN showing comparable performance. 

Surprisingly, across sites, MLP consistently outperforms GCN with the exception of Carinthia. This is different from findings for other traffic-related tasks, such as traffic flow prediction, where, in general, the spatial modeling process of GCNs improves performance \cite{Li2017DiffusionCR}. A reason for this result could be the deployment in low-speed zones or the regulatory speed limit for automated shuttles, leading to less variation in the running times. 

However, we are not only interested in identifying the most competitive model; more importantly, we aim to understand the overall predictability of automated running times and how it varies across different pilot sites. Analyzing $R^2$ scores indicates that, for most cities, simple models like the Lag and Mean Regressors sufficiently explain the variability in running times. With the exception of Madrid, the improvement in prediction accuracy when using more sophisticated models compared to the Mean Regressor (or the Lag model in the case of Les Mureaux) is minimal. Low average speeds and automated driving behavior result in stable running times that simpler models can effectively capture. Madrid, however, stands out with its deployment at a bus depot, focusing on automatic depot management and parking, and therefore showing the lowest average running times (63s) and highest coefficient of variation (0.34). In this case, we can observe a clear benefit of using a prediction model, but even the MAPE and $R^2$ of the best model are significantly higher than for the other sites. The models on Graz show a lower $R^2$ score than in the remaining pilot sites, reflecting data quality issues. Graz has not only the lowest number of observations but, more importantly, the pilot is also scattered across 9 months, leading to data sparsity that leaves the lag feature outdated. 

These findings are further supported by feature importance analysis with SHAP values \cite{NIPS2017_8a20a862} for the XGBoost regressor in Figure \ref{fig:shap_travel}. For all pilot sites except Madrid and Graz, the lag feature is the most influential. In Madrid, the poor predictability of the lag model forces the XGBoost model to rely more on positional data, while in Graz, segment encodings are prioritized due to outdated lags caused by data sparsity. Notably, in Les Mureaux, which operates on a private company site with the lowest average speed, the Lag Model achieves an impressive $R^2$ of 0.95, making the lag feature by far the dominant predictor for this location. This performance underscores the predictive strength of the lag feature under stable, automated conditions.

\begin{table*}
\caption{Results for Dwell Time Prediction}
\label{tab:dwell_time}
\centering
\small
\addtolength{\tabcolsep}{-2pt} 

\begin{tabular}{cccccccccc}
\multicolumn{1}{c}{} & \multicolumn{1}{c}{} & \multicolumn{1}{c}{Lag} & \multicolumn{1}{c}{Mean} & \multicolumn{1}{c}{LR} & \multicolumn{1}{c}{RF} & \multicolumn{1}{c}{XGBoost} & \multicolumn{1}{c}{MLP} & \multicolumn{1}{c}{GCN} & \multicolumn{1}{c}{RF-GCN} \\ \hline
\multirow{3}{*}{Linköping} & RMSE & 15.65           & 12.28                    & 11.43           & 11.14           & \textbf{10.57}  & $11.63 \pm 0.03$ & $10.87 \pm 0.03$         & $10.66 \pm 0.06$ \\
                            & MAE  & 9.15            & \phantom{0}8.78          & \phantom{0}7.63 & \phantom{0}7.03 & \phantom{0}6.59 & $6.92 \pm 0.06$  & $5.92 \pm 0.04$          & $\textbf{5.66} \pm 0.06$ \\ 
                             & $R^2$  & -0.43        & 0.12                     & 0.24            & 0.28             & 0.35           & $0.21 \pm 0.00 $ & $0.31 \pm 0.00$&$\textbf{0.36} \pm 0.00$     \vspace{0.2cm}\\   
                            
\multirow{3}{*}{Tampere}    & RMSE & 17.10           & 13.28                    & 13.07           & 12.69           & \textbf{12.61}  & $13.73 \pm 0.07$ & $13.70 \pm 0.10$         & $12.73 \pm 0.05$ \\
                            & MAE  & 10.57           & \phantom{0}9.73          & \phantom{0}9.46 & \phantom{0}8.75 & \phantom{0}8.55 & $8.20 \pm 0.04$  & $8.13 \pm 0.02$          & $\textbf{7.65} \pm 0.01$ \\ 
                            & $R^2$  & -0.55         & 0.07                      & 0.1           & 0.15             & \textbf{0.16}            & $0.00 \pm 0.01$                     & $0.01 \pm 0.01$                       & $0.14 \pm 0.00$\vspace{0.2cm}\\   
                            
\multirow{3}{*}{Les Mureaux}      & RMSE & 12.46           & 12.26                    & 10.06                       & \textbf{8.06}   & 8.24            & $10.08 \pm 0.13$ & $9.79 \pm 0.18$          & $8.52 \pm 0.06$ \\
                            & MAE  & 5.41            & 8.51                     & 5.85                             & \textbf{3.30}   & 3.46            & $5.45 \pm 0.11$  & $5.23 \pm 0.15$          & $3.98 \pm 0.13$ \\ 
                            & $R^2$  &  -0.13 & -0.09                            & 0.26                            & \textbf{0.53}                             & 0.51                       & $0.26 \pm 0.02$              & $0.30 \pm 0.03$                       & $0.47 \pm 0.01$\vspace{0.2cm}\\   

\multirow{3}{*}{Graz}       & RMSE & 15.84           & 10.88                    & 11.27           & \textbf{10.52}  & 10.66           & $12.71 \pm 0.24$ & $12.92 \pm 0.17$         & $11.21 \pm 0.09$ \\
                            & MAE  & 11.59           & \phantom{0}\textbf{7.13} & \phantom{0}7.16 & \phantom{0}7.89 & \phantom{0}7.94 & $8.76 \pm 0.27$  & $8.77 \pm 0.10$          & $7.24 \pm 0.08$ \\ 
                            & $R^2$  &  -1.12 & 0.00                            & -0.07                           & \textbf{0.07}                             & 0.04                                    & $-0.37 \pm 0.05$                      &$ -0.41\pm -0.04$  & $-0.06\pm 0.02$\vspace{0.2cm}\\   
                            
\multirow{3}{*}{Madrid}     & RMSE & 22.83           & 19.41                    & 18.49           & 18.03           & \textbf{16.19}  & $16.90 \pm 0.17$ & $17.06 \pm 0.2$          & $17.18 \pm 0.09$ \\
                            & MAE  & \phantom{0}6.58 & \phantom{0}9.71          & \phantom{0}9.27 & \phantom{0}7.77 & \phantom{0}6.20 & $4.51 \pm 0.38$  & $\textbf{4.21} \pm 0.01$ & $4.62 \pm 0.10$ \\ 
                            & $R^2$  &  -0.90 & -0.37                            & -0.24                           &  -0.18                             & \textbf{0.05}                                    & $-0.04 \pm 0.02$                       & $-0.06 \pm 0.00$& $-0.07 \pm 0.01$ \vspace{0.2cm}\\   
                            
\multirow{3}{*}{Carinthia}  & RMSE & 23.70          & 19.08                   & 18.26            & 17.34   & \textbf{17.31}          & $18.90 \pm 0.17$ & $19.44 \pm 0.40$          & $18.12 \pm 0.02$ \\
                            & MAE  & 14.92          & 13.93                   & 12.87            & 12.03   & 11.76                    & $12.23 \pm 0.08$  & $12.56 \pm 0.54$         & $\textbf{11.05} \pm 0.02$ \\ 
                            & $R^2$  &  -0.21 & 0.22                            & 0.28                            & 0.35                             & \textbf{0.36}                                    & $0.23 \pm 0.01$ &$0.18 \pm 0.03$ & $0.29 \pm 0.00$\\   
\end{tabular}%
\end{table*}

\begin{figure}
    \centering
    \includegraphics[width=\linewidth]{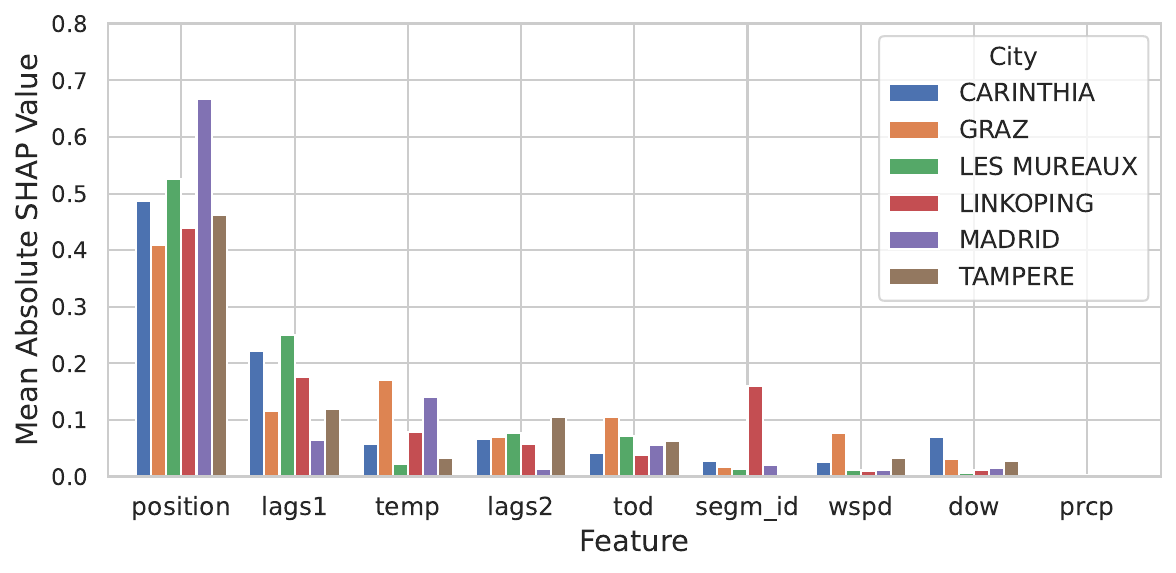}
    \caption{Mean SHAP values for dwell time prediction with XGB.} 
    \label{fig:shap_dwell}
\end{figure}

\subsection{Dwell time prediction}
In Table \ref{tab:dwell_time}, we analyze whether the results from running time predictions translate to dwell time predictions. Interestingly, there is no universal best model across all sites, showing quite a different picture from Table \ref{tab:travel_time}. This is not unexpected, as running time and dwell time have different characteristics. In Linköping, Tampere, and Carinthia, XGBoost outperforms in RMSE, while RF-GCN achieves the lowest MAE, with the difference in MAE being more significant than in RMSE. These results highlight the benefit of integrating spatial information for dwell time prediction in suburban residential areas, where these three pilots are situated. 
In Les Mureaux, operating employee shuttles, the situation is reversed. The tree-based models outperform the other models by a large margin. We perform a more detailed analysis of this difference in Section \ref{sec:dwell2}, identifying automated dwell times as a contributing factor. Given the very limited data availability in Graz, it is no surprise that our tested models cannot outperform the Mean Regressor, as there is not enough data for the advanced models to learn the spatial and temporal correlations. In Madrid, the GCN demonstrates better performance than RF-GCN. We can attribute this to an observed degradation in the performance of the RF classifier in detecting zeros compared to the standalone GCN. This intriguing finding suggests that spatial information in this pilot, which does not focus solely on passenger transportation, not only helps in accurate dwell time prediction during stops but also in detecting stop skips.

Again, we are not only interested in the best-performing prediction models but also in understanding the general predictability of dwell times for automated shuttles. Unlike in running time prediction, both baseline models, Lag and Mean, yield low and sometimes even negative $R^2$ values across all pilot sites, highlighting dwell time prediction as the more challenging task. Given these results, it is surprising that most literature focuses on travel or running time prediction, while dwell time prediction has received far less attention. 
The fact that the lag feature is less informative for dwell time prediction than for running time prediction is also reflected in the feature importance analysis in Figure \ref{fig:shap_dwell}. Here, position stands out as the most influential feature by a large margin. As a result, the model must rely on a broader range of input features for accurate dwell time prediction, i.e., temperature and time-of-day features, which only had minimal impact on running time.

Evaluating the $R^2$ of the best-performing models reveals substantial differences across pilot sites, as expected given their different characteristics. In Graz and Madrid, the models barely achieve positive $R^2$ values. In Graz, this can largely be attributed to the limited data availability. Additionally, as the only pilot site using passenger cars, factors such as low-clearance boarding or alighting, along with passengers manually operating the doors, may cause dwell times to be more dependent on passenger characteristics (e.g., mobility) and the number of passengers (e.g., when the exit is limited to a single door). This situation seems to lead to harder-to-predict dwell times compared to conventional shuttle buses. 
These findings are supported by the results from Tampere, which show a lower $R^2$ score than Linköping or Carinthia, despite operating in areas with similar characteristics and having a comparable or higher number of observations but deploying a van instead of a bus. The poor results in Madrid are attributed to the focus on automated depot management, resulting in very high variability in dwell times as noted in the previous section. This variability appears to be driven more by the testing of operational procedures than by consistent passenger demand patterns, leading to greater unpredictability. Finally, in Les Mureaux, $R^2$ scores are significantly higher than in the other sites, which can be explained by an automated, default dwell time at this pilot site, a factor that will be explored in more detail in the next section. Overall, across all sites except Graz, using ML-based models significantly improves performance, demonstrating their value in dwell time predictions.

\subsection{Detailed analysis of dwell time predictions}
\label{sec:dwell2}
The previous section showed that the optimal choice between deep learning and tree-based methods for predicting dwell times is site-dependent, with no universally best model. In light of this, a deeper understanding is needed to determine the conditions that favor one approach over the other.

\begin{figure*}
    \centering
    \begin{subfigure}{.5\linewidth}
        \centering
        \includegraphics[width=\linewidth]{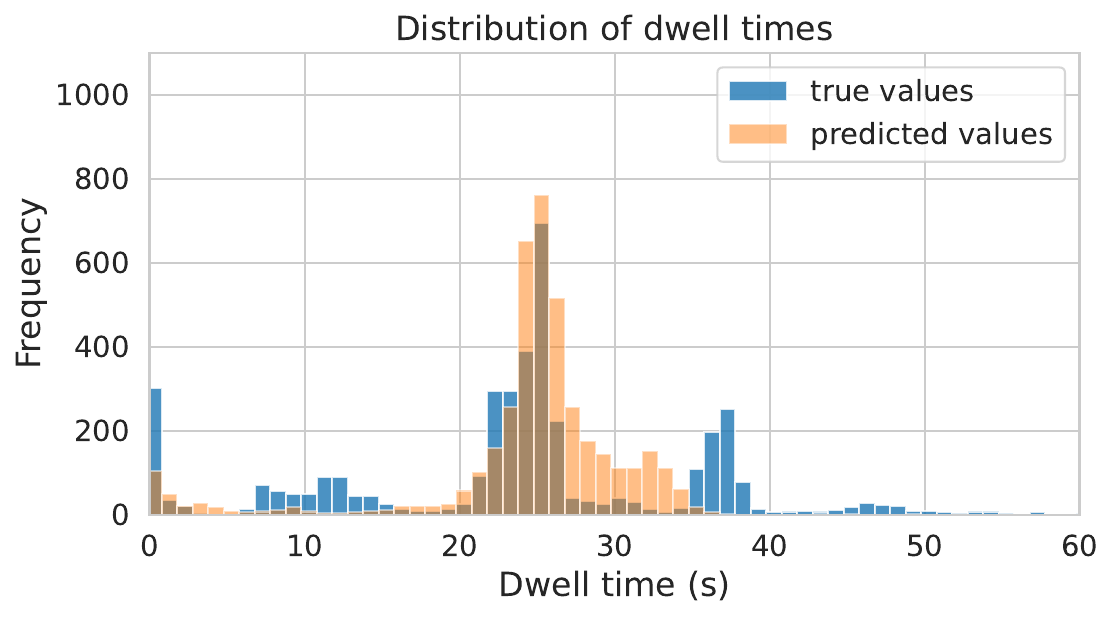}
        \caption{XGBoost Predictions} 
        \label{fig:lin_dwell1}
    \end{subfigure}%
    \begin{subfigure}{.5\linewidth}
        \centering
        \includegraphics[width=\linewidth]{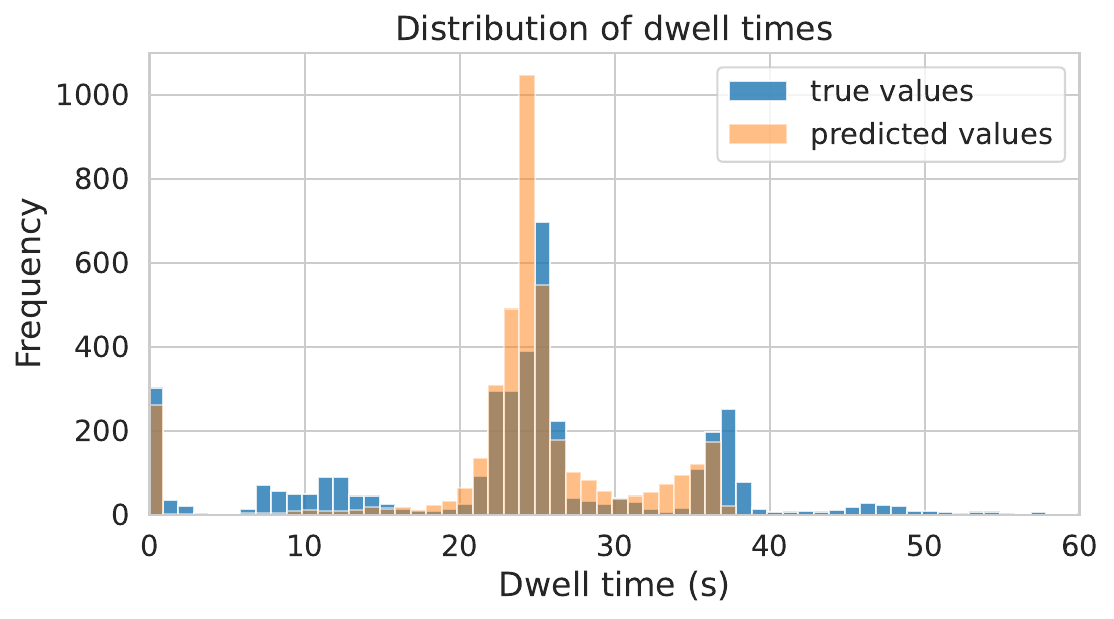}
        \caption{RF-GCN Predictions} 
        \label{fig:lin_dwell2}
    \end{subfigure}
    \caption{Dwell time Distribution Linköping}
    \label{fig:dwell_linkoping}
\end{figure*}

\begin{figure*}
    \centering
    \begin{subfigure}{.5\linewidth}
        \centering
        \includegraphics[width=\linewidth]{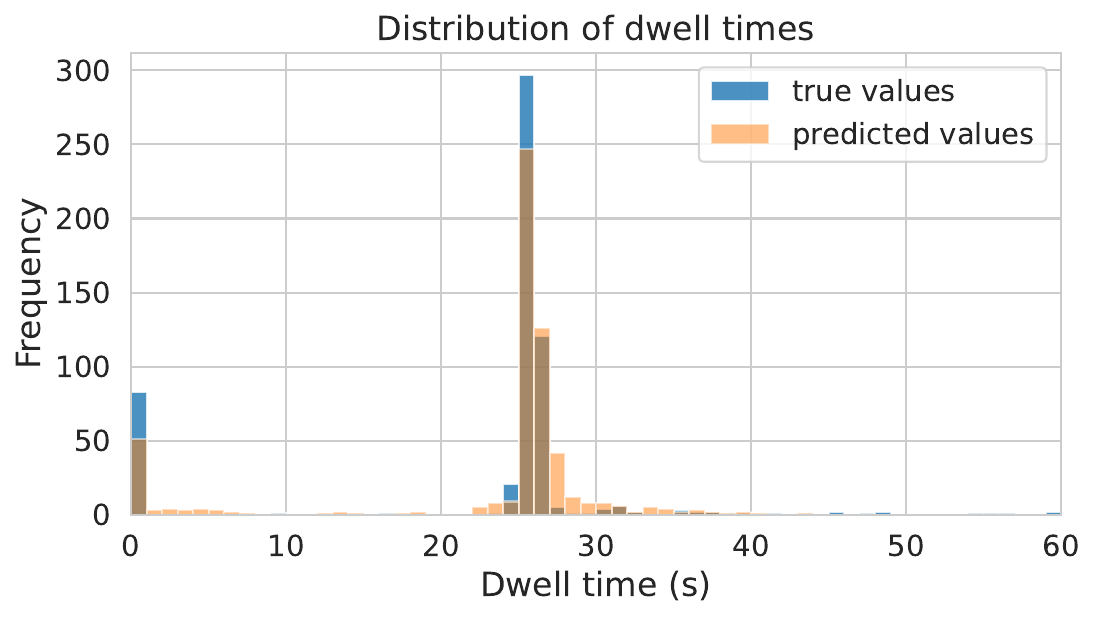}
        \caption{Random Forest Predictions} 
        \label{subfig:image1}
    \end{subfigure}%
    \begin{subfigure}{.5\linewidth}
        \centering
        \includegraphics[width=\linewidth]{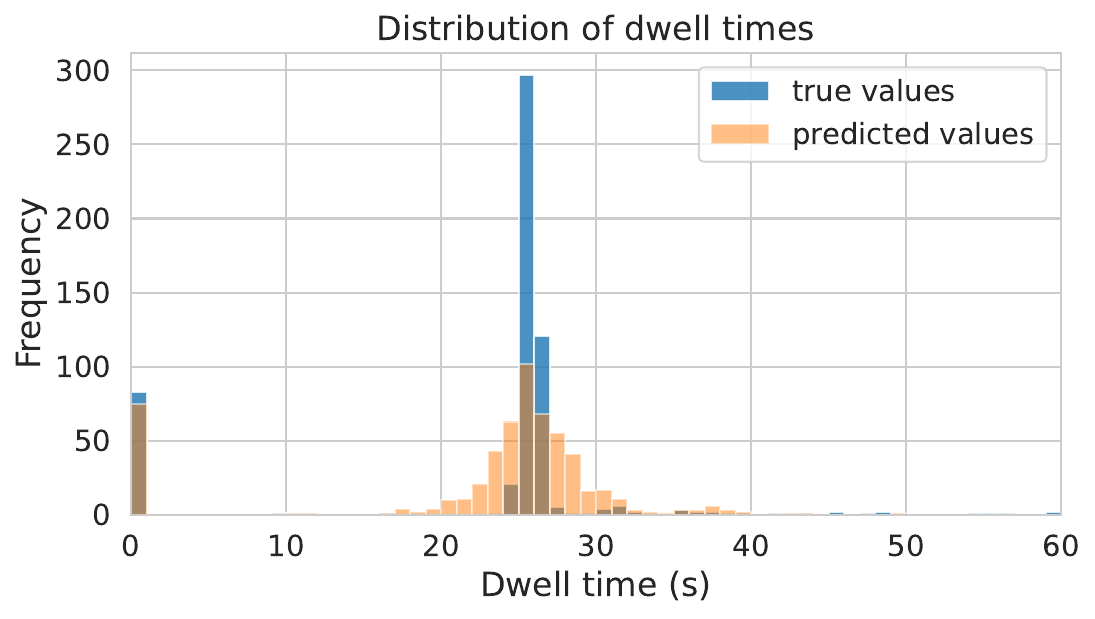}
        \caption{RF-GCN Predictions} 
        \label{subfig:image2}
    \end{subfigure}
    \caption{Dwell time Distribution Les Mureaux}
    \label{fig:dwell_rouen}
\end{figure*}

In Figure \ref{fig:dwell_linkoping}, we depict the dwell time distribution in Linköping. Linköping is unique in having the longest ongoing pilot and route, collecting nearly five times more observations and serving more than double the number of passengers as the other sites. The dwell time distribution here is distinctly multimodal. A mode at zero seconds shows instances where a shuttle stop is skipped. The remaining modes suggest times when the shuttle stops and passengers are alighting or boarding. There is a variation of the dwell times between 20 and 40 seconds, indicative of a frequently used service. In Figure \ref{fig:dwell_linkoping}, we observe a difference in the performance of the prediction models. Figure \ref{fig:lin_dwell1} illustrates that XGBoost struggles to categorize the majority of zero dwell times and overlooks the third mode at 35 seconds. In contrast, our RF-GCN (\ref{fig:lin_dwell2}) demonstrates a better fit. The RF classifier effectively classifies whether a shuttle is skipping a stop, and the GCN model fits the multimodal dwell time distribution for cases when passengers are alighting and boarding. Remarkably, by incorporating spatial information, this model even differentiates between the two primary dwell times of approximately 25 and 35 seconds. 

On the other hand, the benefits of tree-based methods are evident in the case of Les Mureaux (Figure \ref{fig:dwell_rouen}), where the shuttles are only deployed for employees on a private company site and thus shows the second lowest number of served passengers. Here, we can notice the automated nature of the service with two main dwell times of 0 and 25 seconds for skipping and stopping, respectively. The data from this site suggests a default dwell time of 25 seconds that extends only when necessary. As prior research has shown \cite{Grinsztajn}, neural networks are biased towards overly smooth solutions. In contrast, decision trees, learning piece-wise constant functions, avoid this pitfall. This characteristic is evident in Figure \ref{fig:dwell_rouen}, where we can observe the superiority of the Random Forest when fitting to a non-smooth target function. 

Therefore, we conclude that the relative advantage of using a GNN depends on the specific characteristics of the shuttle service. In cases with high data availability, the frequently-used service with a long route, and deployment in suburban, residential areas,  the GNN seems superior. Yet, the random forest remains relevant in detecting the skipped stops. However, in scenarios where the dwell times are largely automated, do not significantly vary with passenger density, or service types with lower passenger counts (e.g., employee-only shuttles), the prediction challenge simplifies to a classification task: is the shuttle stopping or not? In these conditions, the tree-based methods show a clear advantage. 

\subsection{Evaluation on journey}

\begin{figure*}
    \centering
    \begin{minipage}{.5\linewidth}
        \centering
        \includegraphics[width=0.95\linewidth]{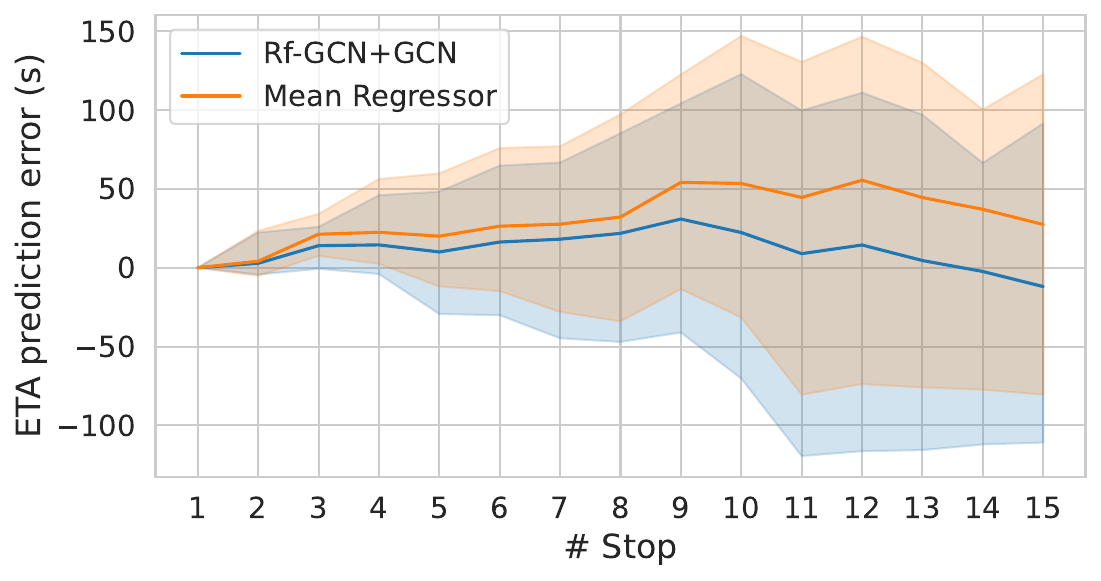}
    \end{minipage}%
    \begin{minipage}{.5\linewidth}
        \centering
        \includegraphics[width=0.95\linewidth]{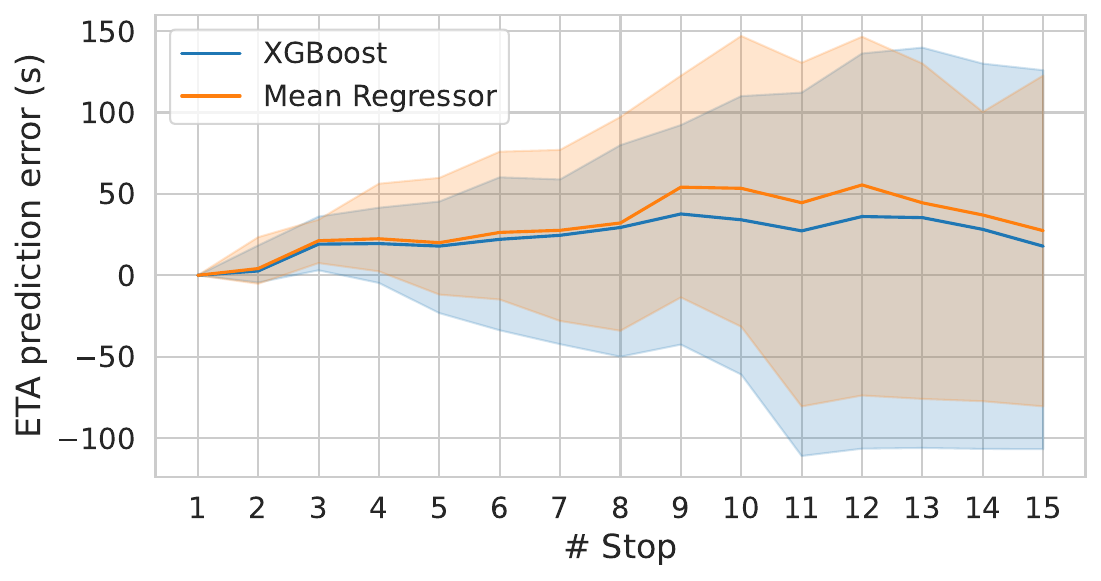}
    \end{minipage}
    \caption{AT Prediction Linköping}
    \label{fig:eta_linkoping}
\end{figure*}

\begin{figure*}
    \centering
    \begin{minipage}{.5\linewidth}
        \centering
        \includegraphics[width=0.95\linewidth]{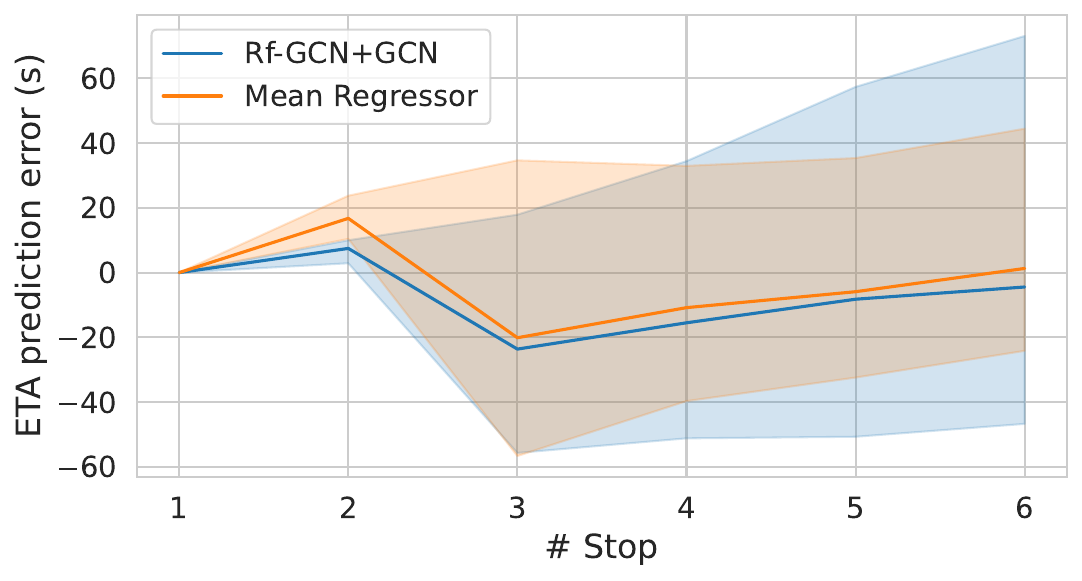}
    \end{minipage}%
    \begin{minipage}{.5\linewidth}
        \centering
        \includegraphics[width=0.95\linewidth]{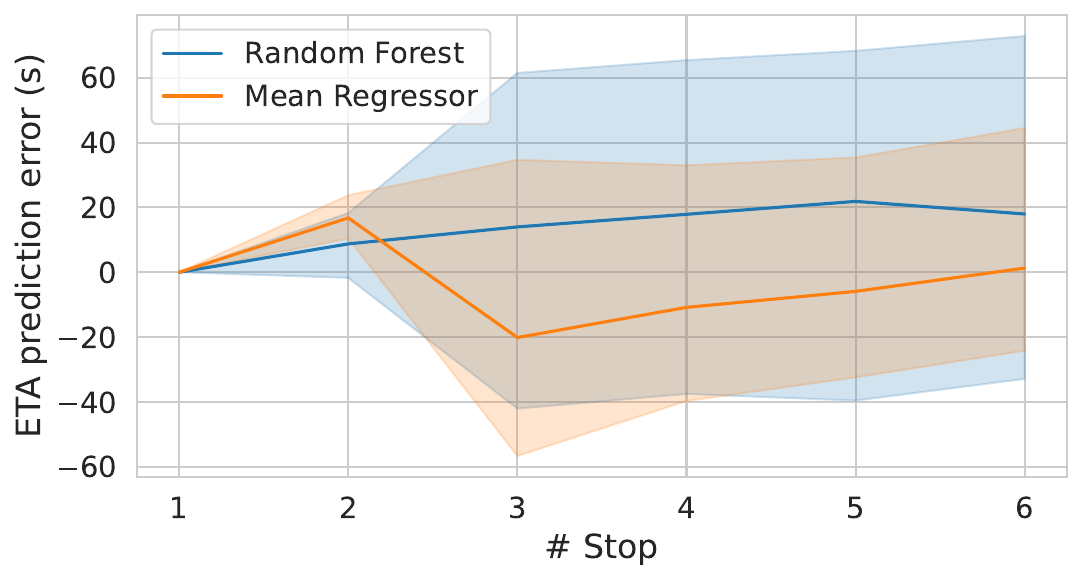}
    \end{minipage}
    \caption{AT Prediction Les Mureaux}
    \label{fig:eta_rouen}
\end{figure*}

\begin{figure*}
    \centering
    \begin{minipage}{.5\linewidth}
        \centering
        \includegraphics[width=0.95\linewidth]{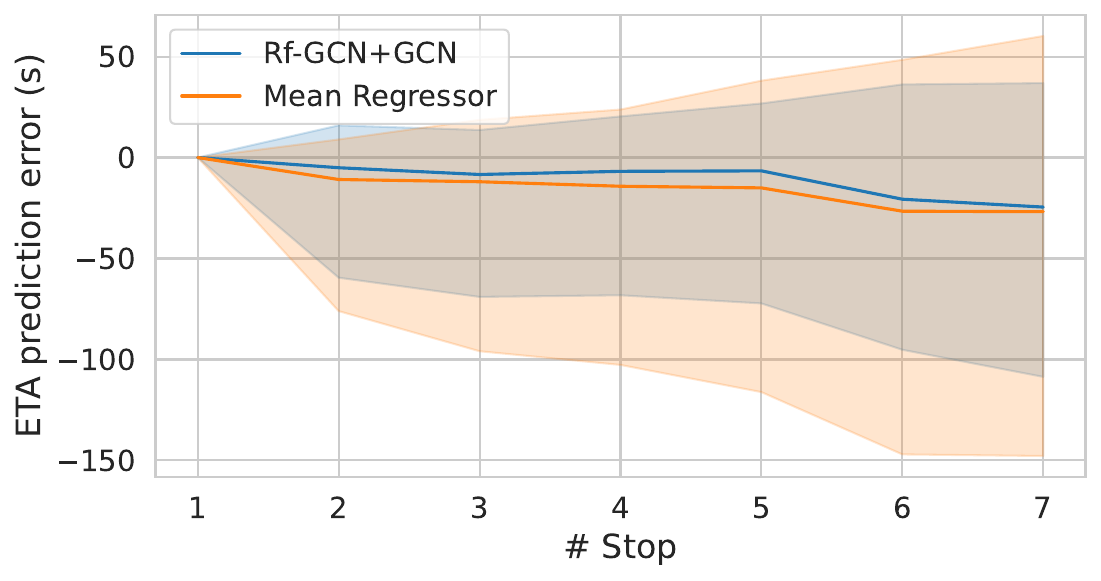}
    \end{minipage}%
    \begin{minipage}{.5\linewidth}
        \centering
        \includegraphics[width=0.95\linewidth]{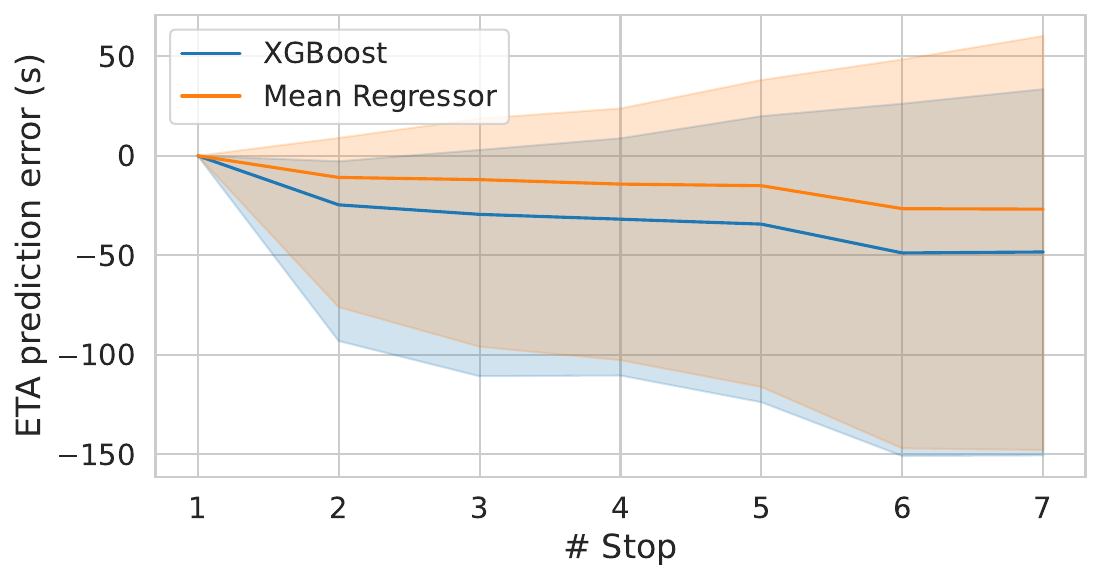}
    \end{minipage}
    \caption{AT Prediction Tampere}
    \label{fig:eta_tampere}
\end{figure*}

So far, our focus has been on evaluating model performance over individual segments. However, in real-world service deployment, the cumulative AT prediction, formed by aggregating individual segment predictions, is more important. Hence, we sample five journeys from the test set and assess the accumulated error along the journey starting from stop number 1. We illustrate the results for the three sites with the longest routes, Linköping, Les Mureaux, and Tampere, in Figures \ref{fig:eta_linkoping}-\ref{fig:eta_tampere}. Solid lines depict average accumulated errors, while the shaded regions represent the range between maximum positive and negative errors across the sampled journeys. At each prediction instant, we freeze the features to whatever data is available while the shuttle is still at stop 1. Lag variables, therefore, reflect the most recent observations before the current trip begins; they are not updated for predictions further ahead. We evaluate the accumulated error across sample strips to evaluate systemic prediction bias and compounding errors.

Linköping has the longest route of the pilot sites, with a total of 15 stops. In Figure \ref{fig:eta_linkoping}, stop number 7 corresponds to a 15-minute, and stop number 15 to a 35-minute ahead prediction. The depicted models (RF-GCN, XGBoost, Mean Regressor) overestimate the total travel time, but RF-GCN shows the least average error with a maximum of 30 seconds. In the 15-minute ahead predictions at stop number 7, we can observe an error of a maximum of 60 seconds and 35 minutes ahead of around 100 seconds. XGBoost still shows a lower average error than the Mean Regressor, yet, toward the end of the route, the accumulated errors display worse maximum values.

In Les Mureaux (Figure \ref{fig:eta_rouen}), the full journey is shorter, spanning six stops, which is equivalent to a 20-minute travel time. Here we observe that the accumulated errors do not exceed 70 seconds in over- and 60 seconds in underestimating. 

The interesting finding in the Tampere (Figure \ref{fig:eta_tampere}) site is that although XGboost outperforms the GCN on running time prediction, the accumulated predictions by RF-GCN/GCN are more accurate. The average error is close to zero, and the maximum error stays below 100 seconds, while XGBoost is underestimating the running time by up to 50 seconds on average and 150 seconds in the worst case. These findings suggest that future work needs to rethink model evaluation, as most existing works do not consider aggregation errors.

\begin{table*}
    \caption{Averaged Accumulated Absolute Error (s)} \label{tab:accumulated_error}
    \begin{center}
    \small
    \addtolength{\tabcolsep}{-2pt}
    \begin{tabular}{l  c c c c c c}
        & \multicolumn{2}{c}{Linköping} & \multicolumn{2}{c}{Tampere} & \multicolumn{2}{c}{Les Mureaux}\\
Model   & Dwell Time & Running time         &  Dwell Time & Running time      & Dwell Time & Running time    \\ \hline
Mean    & 142        & 126              & 76          &61             & 36         & 72                      \\
XGB/RF  & 106        & \textbf{105}     & 73          &\textbf{52}    & \textbf{14} & \textbf{49}          \\   
GCN/GCN-RF     & \textbf{93} & \textbf{105}    & \textbf{56} &57             & 16         & 64                \\ \hline
		\end{tabular}
	\end{center}
\end{table*}

Since we are aggregating the predictions of the dwell and running time prediction models, the question arises of how the errors accumulate between those two models and which one is driving the difference in performance in Figures \ref{fig:eta_linkoping}-\ref{fig:eta_tampere}. In Table \ref{tab:accumulated_error}, we calculate the average accumulated absolute error over the entire journey resulting from the dwell and the running time prediction, respectively, which is calculated by $\frac{1}{ |J |}\;\sum_{j=1}^{J}\;\sum_{k=1}^{N_j}\Bigl|{\hat{s}_{k}}\;-\;s_{k}\Bigr|$, where $J$ is the set of Journeys and $N_j$ is the set of segments or stops in journey $j$, $\hat{s}_{k}$ the predicted time for segment $k$, and $s_{k}$ the observed time for segment $k$. The performance of the Mean Regressor on Linköping and Tampere reveals that the error originating from the dwell time prediction has a larger share of the total travel time error than the one from the running time model. Similarly, we can identify the reason for the superior performance of the GCN in the accumulated travel times, as it decreases the error in dwell time prediction significantly while also further decreasing the error in running time prediction. 

Interestingly, while evaluating individual segments in Section \ref{sec:run_pred} did not show a clear advantage for using a prediction model for running time, the accumulated errors reveal a notable improvement when the errors are accumulated over the entire journey. More specifically, if we compare the decrease of errors of the best model to the Mean Regressor, we observe a decrease of 35\%, 26\%, and 61\% in dwell time error for the three pilot sites and a decrease of 17\%, 15\%, and 32\% for the error originating from the running time predictions. With these results, we stress that dwell time prediction is an important factor for the AT prediction for automated shuttles. Especially when they are used in remote, low-speed areas or are confined to regulatory speed limits, dwell time becomes the most influential factor.

\begin{figure}
    \centering
    \includegraphics[width=\linewidth]{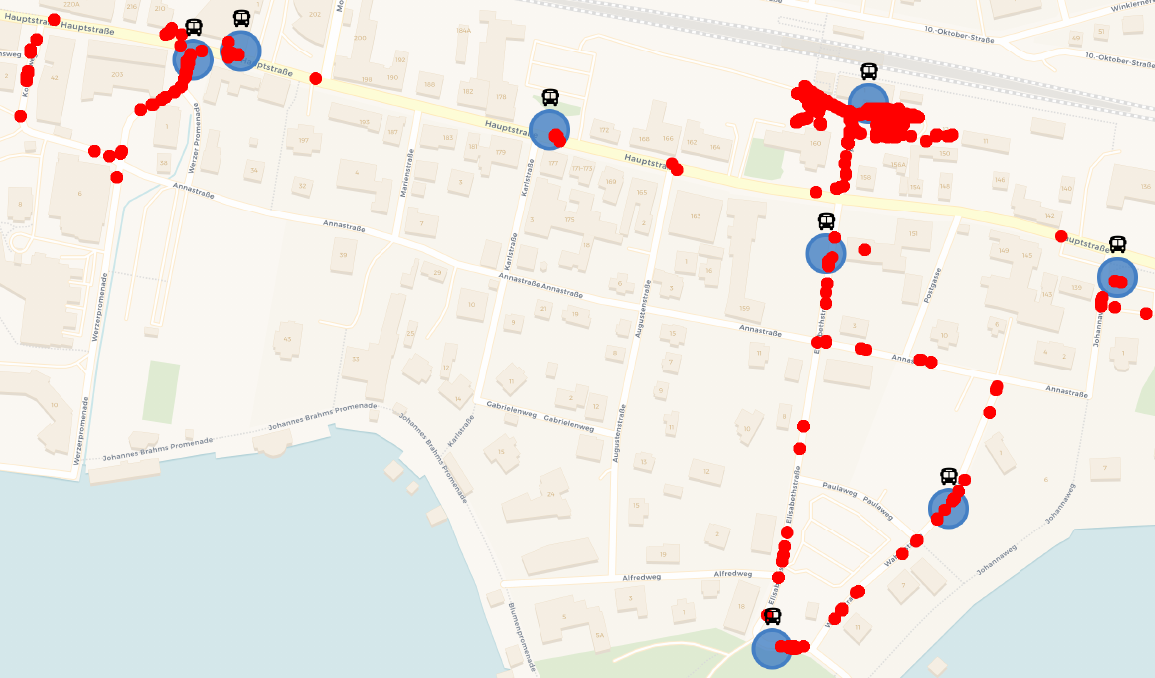}
    \caption{Stop location in the Carinthia pilot site (blue) and locations with opened door (red)} 
    \label{fig:carinthia_door}
\end{figure}

\subsection{Deviation from scheduled, automated service}
\begin{figure}
    \centering
    \includegraphics[width=\linewidth]{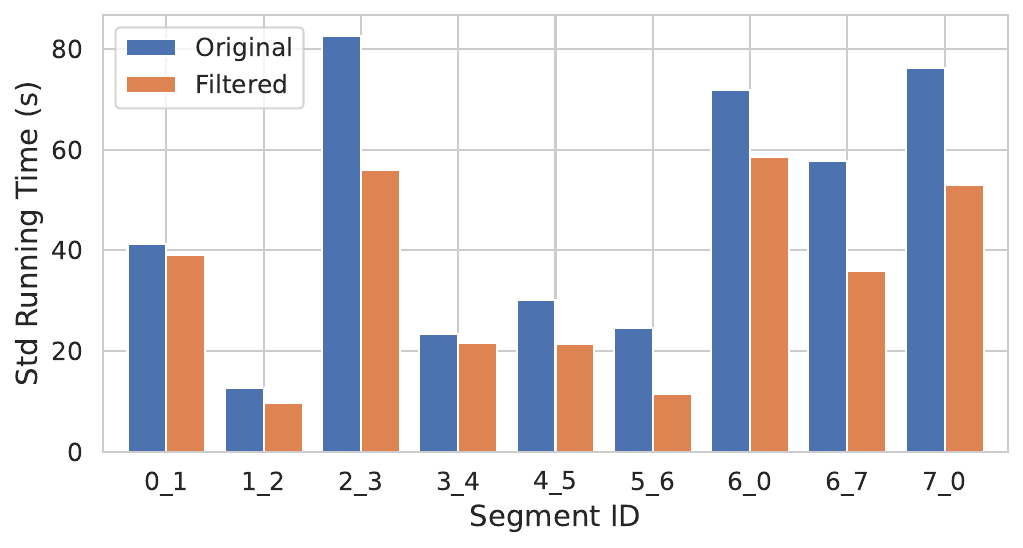}
    \caption{Comparison of standard deviation in running times when filtering out on-demand service} 
    \label{fig:carinthia_std}
\end{figure}

Some pilot sites have reported allowing safety drivers to make unscheduled stops to let passengers board in an effort to increase public acceptance. This manual on-demand service, being unplanned, is not recorded in the datasets. For one of the pilot sites, Carinthia, we, however, do have access to door status information, which allows us to track stops outside the scheduled service. Figure \ref{fig:carinthia_door} shows the scheduled stops alongside all door-opening occurrences, highlighting the challenges of working with real-world data. Our arrival time prediction approach is designed for scheduled, automated shuttle services, and these unplanned stops introduce variability in running times that is not captured by our features, such as time of day or weather conditions, but rather depends on the decisions of the safety driver. Specifically, we observe that unplanned stops occurred during $20\%$ of the running time segments, adding noise to the overall running time calculation and increasing complexity for prediction models. To provide further insight, Figure \ref{fig:carinthia_std} shows the reduction in variability (i.e., in standard deviation) by filtering out the segments with stops en-route. For longer segments, such as \textit{2\_3} or \textit{6\_7}, unplanned stops account for more than $30\%$ of the running time variability rather than the actual driving time. 

This variability impacts prediction model performance. By evaluating on the filtered dataset, we observe a decrease in MAE of $1.8\%$ for the Mean Regressor, $4.5\%$ for XGBoost, and $9.2\%$ for the Random Forest regressor. These results show that machine learning methods are more susceptible to the additional noise introduced by unplanned stops, highlighting the importance of filtering or accounting for such variability to improve prediction accuracy.

\subsection{Key insights from working with data from real-world automated shuttle deployment}
Data gathered during pilot tests often comes with limitations. Most sites provide only GPS positioning and speed data. Given this constraint, we cannot rely on many of the input features that have been referenced in prior literature, such as passenger data, which have been proven to be the most impactful for dwell time prediction \cite{RASHIDI202332}. It is reasonable to assume that demand is highly volatile during pilot phases, as some passengers might be merely curiosity-driven, keen on experiencing technological advancement just once. 
This variability further complicates dwell time predictions. Given that the dwell time of zero emerged as a challenge in our prediction models and the error in dwell time prediction had more impact on the overall AT prediction, passenger monitoring promises a great influence on the final AT prediction. 


Regarding running time prediction, we also encounter challenges. As automated shuttles operate on both shared public roads and private lanes or share lanes only with pedestrians and cyclists, we cannot draw from standard road traffic speed estimates available from car data. Instead, we are limited to the most recent running time observations from the shuttle. In some test scenarios where only a single vehicle was operational, or operations were stopped due to technical issues, these recent observations can quickly become outdated and, thus, less reliable. Further, local regulations constrain the speed limit of the shuttle, limiting the variance in running times and, therefore, the potential improvement obtained by predictions. 

In the previous section, we quantified the impact of deviations from scheduled operations that were not documented in the data. Thus, our key recommendation for practitioners collecting data during pilots is to be mindful of data consistency. If ordinary operations are interrupted, data collection should either be paused or the collected data should be labeled as 'non-ordinary'. Without this information, we cannot guarantee that test runs to check technical feasibility without the intention of transporting passengers are not included in the data set. Similarly, given that the vehicles are automated at level 4 with a safety driver who may interfere with operations— such as letting passengers board between stops— it is essential to flag any instances of safety driver intervention. Failing to do so could undermine the validity of conclusions drawn about level 5 automation. Furthermore, any changes to the settings of the shuttle, such as dwelling settings, the sampling rate of the GPS device, or acceleration/deceleration behavior, should be documented. This would greatly improve data processing and the evaluation of predictions for regular operations post-pilot. The conclusions drawn in this study are scoped to the current deployment context of automated shuttles, i.e., smaller services that complement existing public transport. We deliberately do not generalize to high-capacity trunk services on primary roads, where higher speed limits and traffic volumes may change the value of running time prediction, warranting a dedicated study. As technology and deployments evolve, model choices should be re-evaluated.

\section{Conclusion}
This work studied the impact of automated vehicles on prediction models within public transport. As autonomous transportation emerges, accurate AT predictions are vital to gaining public acceptance. Given the early stage of this technology, models typically have to be applied to data originating from pilot tests with automation level 4 (with a safety driver). Hence, we emphasized the importance of careful data collection to practitioners to obtain an estimate of the prediction for regular operations post-pilot. 

We tested several established machine learning and deep learning models on dwell and running time prediction in the setting of automated mobility using data from six diverse pilots across Europe. A key finding is the absence of a universally best model for dwell time prediction. In cases with high data availability, a long, frequently-used route, and deployment in suburban residential areas, the hierarchical approach of a random forest classifier combined with a graph neural network is superior. However, in scenarios where the dwell time is mainly automated, does not significantly vary with passenger density, and the service type has lower passenger counts (e.g., employee-only shuttles), the prediction challenge simplifies to a classification task, and the tree-based models are preferable. 


We found that dwell time predictability declines when (i) the pilot site’s focus extends beyond passenger transportation, (ii) passenger cars or vans with manual doors and low-clearance boarding are used instead of shuttle buses, and (iii) data availability is limited. The greatest improvement in predictability occurs with automated dwell times, where a standard door opening time is applied, underlining the critical role of dwell time prediction in accurate AT estimation for automated shuttles. 
Overall, advanced prediction models showed a clear advantage over simpler approaches like the Mean or Lag model, which failed to capture data variability. 

For running time prediction, however, the benefit of advanced prediction models was less clear. Surprisingly, leveraging deep learning architectures that incorporate spatial information failed to yield clear benefits over tree-based methods. This observation can be partly attributed to the fact that the shuttles are mainly deployed in low-speed zones or are constrained by a regulatory speed limit for automated shuttles. 
Consequently, for automated shuttles at their current stage of development, the Mean or Lag Regressor already provides adequate explainability for most pilot sites when evaluating individual predictions. However, when the goal is to provide accurate predictions well in advance, a prediction model can help reduce the accumulated error. 

To summarize, at the current stage of automated shuttle deployment, we observed contrasting trends in dwell and running time predictions, with advanced models proving more valuable for dwell and less critical for running time prediction. Consequently, for the final AT prediction, dwell time prediction proves to be the driving component, with a critical model capability being able to detect skipped stops. Encouragingly, the results for the final AT prediction are promising, with maximum errors of 60 seconds, equaling maximum absolute percentage errors of $15\%$ for predictions up to five stops ahead.  

Most importantly, our findings are not consistent with the main focus of most of the literature, which develops sophisticated (GNN) models for running and travel time prediction. In contrast, dwell time prediction —shown to be especially influential for automated shuttles— has received comparatively less attention and typically relies on simpler models. Additionally, in working with noisy real-world data, our work suggests that factors such as careful data collection, thorough data preparation, and the consideration of operational models are more impactful than the final choice of prediction model. 
 
Future research directions are threefold. First, as automated public transport systems evolve in complexity, research should monitor performance changes in predictions in response to growing passenger demand and increased speed limits. Secondly, there is room to explore other models, such as combining the hierarchical zero-inflated approach into one model with combined optimization, e.g., with a Gaussian mixture model. Lastly, future research should collect and incorporate additional data features into the prediction models, i.e., on-board sensor data. As such, compiling a consistent POI/land-use layer is a valuable next step, and combining such static demand proxies with future automatic passenger-count data would likely yield the greatest improvement in dwell
time prediction. Specifically, passenger data, along with details on shuttle characteristics and operational settings that affect dwell time, acceleration, and cruising speed, are potentially important data characteristics that influence the AT prediction of automated shuttles.

\appendices

\appendix
\begin{table}[!h]
\small
\caption{Deployment statistics by site.}
\centering
\setlength{\tabcolsep}{4pt}

\begin{subtable}[t]{\columnwidth}
\centering
\caption{Network geometry. Network density equals to the total length of drivable roads.}
\begin{tabular}{l S[table-format=1.2] S[table-format=2.2] S[table-format=3.0]}
\toprule
Site &
\multicolumn{1}{c}{\shortstack{Service Area\\(\si{\kilo\meter\squared})}} &
\multicolumn{1}{c}{\shortstack{Network density\\(km/km$^2$)}} &
\multicolumn{1}{c}{\shortstack{Intersections}} \\
\midrule
Linköping   & 0.94 & 75.00 & 76 \\
Tampere     & 0.59 & 81.54 & 123 \\
Les Mureaux & 0.31 & 54.79 & 14 \\
Madrid      & 0.05 & 43.65 & 9 \\
Graz        & 0.30 & 55.33 & 29 \\
Carinthia   & 0.23 & 35.61 & 22 \\
\bottomrule
\end{tabular}
\end{subtable}
\vspace{0.45em}

\begin{subtable}[t]{\columnwidth}
\centering
\caption{POIs and road-type mix. POIs include amenity, shop, leisure, tourism, office; “POIs @200m” are POIs within 200\,m of each stop (median [IQR]). Road types are length-weighted shares of top classes.}
\begin{tabularx}{\columnwidth}{l S[table-format=3.0] l X}
\toprule
Site & {POIs} & {POIs@200\,m} & {Top road types (share \%)} \\
\midrule
Linköping   & 172 & 27\,[17–50] & service (39), living (20), \phantom{00} residential (16) \\
Tampere     & 235 & 19\,[13–30] & service (65), residential (20), secondary (8) \\
Les Mureaux & 11  & 1\,[0–4]    & service (100) \\
Madrid      & 3   & 0\,[0–1]    & service (100) \\
Graz        & 51  & 14\,[6–27]  & service (66), residential (27), primary (5) \\
Carinthia   & 107 & 35\,[26–46] & residential (66), service (19), primary (15) \\
\bottomrule
\end{tabularx}
\end{subtable}
\label{tab:osm}
\end{table}
We provide a context table derived from OpenStreetMap \cite{OpenStreetMap} (Table~\ref{tab:osm}), computed within each service area. For each site, we report service area ($km^2$), drivable road density ($km/km^2$) as a measure of network density, consolidated intersection count (3‑way+), total POIs, POIs within 200 m per stop (median [IQR]), and the length‑weighted road‑type mix (e.g., service, primary, residential). These proxies characterize deployment context and complement Tables~\ref{tab:show1}-\ref{tab:show2} by quantifying network and demand features. \footnote{Because Open Street Map coverage varies by location, values should be interpreted as context descriptors rather than exhaustive counts.}

Table \ref{tab:osm} underlines the deployment context of automated shuttles as smaller services deployed in residential or service (retail/campus/company) areas. 
Service-dominated sites such as Les Mureaux have sparse intersections and low POI proximity, conditions under which tree-based models perform strongly. By contrast, mixed-traffic settings (e.g., Linköping, Tampere, Carinthia) exhibit richer connectivity and amenity density, and benefit from spatial models once observation counts are moderate.  

\ifCLASSOPTIONcaptionsoff
  \newpage
\fi



\bibliographystyle{IEEEtranN}
\bibliography{my_bib}
%

%
\vspace{-3em}
\begin{IEEEbiography}
[{\includegraphics[width=1in,height=1.25in, clip,keepaspectratio]{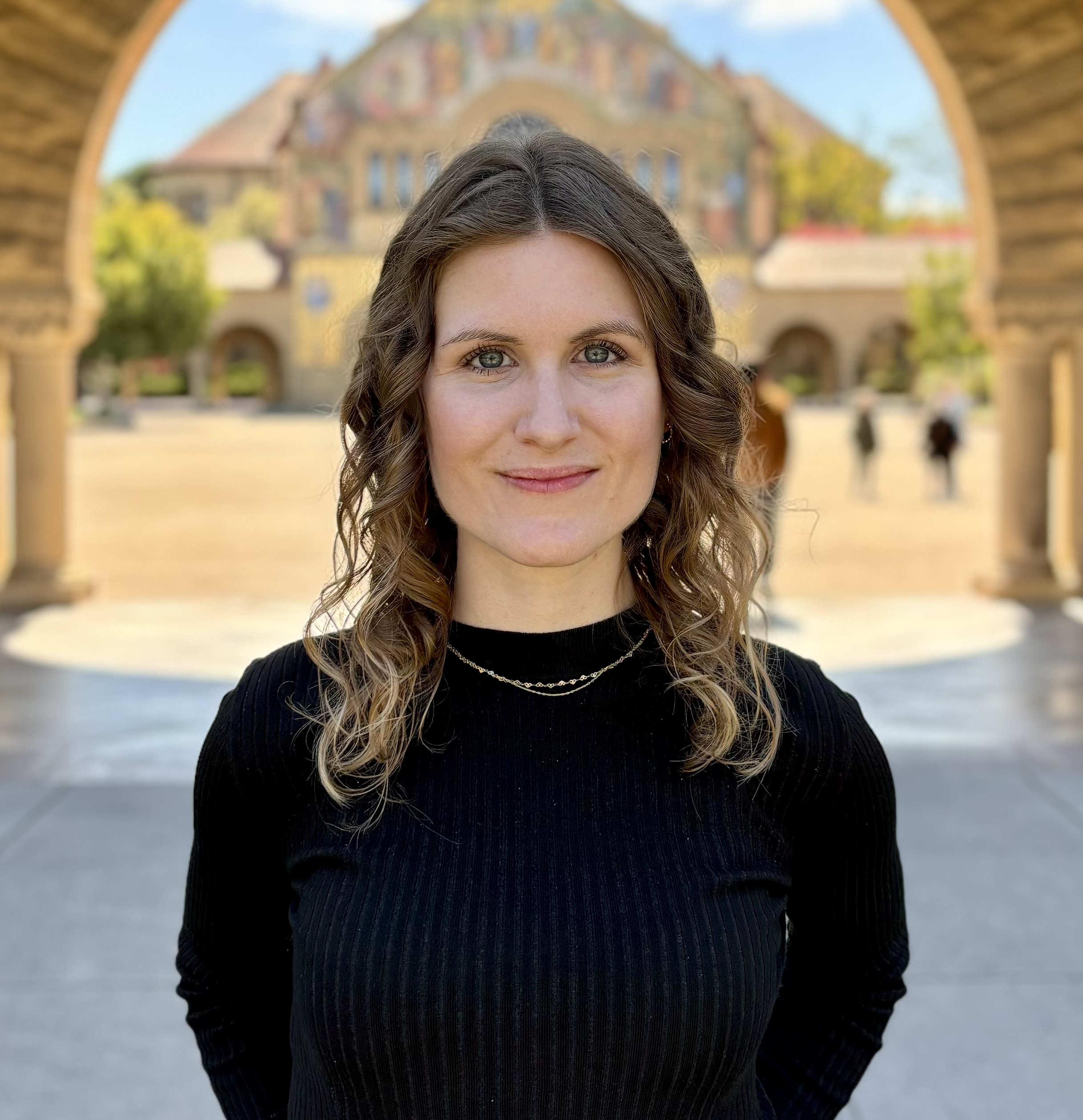}}]{Carolin Schmidt} is a Postdoctoral Scholar at the Technical University of Denmark in the Machine Learning for Smart Mobility lab, where her research includes learning-based control for autonomous mobility systems and climate change adaptation. She completed her Master's degree in Electrical Power Engineering and Operations Research with Honors at RWTH Aachen University. As part of her PhD, she was involved in an EU-wide project on the real-world deployment of autonomous shuttles. During her time as a visiting researcher at the Autonomous Systems Lab at Stanford University, she worked on integrating deep reinforcement learning (RL) with established optimization and control methods to create scalable, robust, and safe RL, contributing to applications in large-scale transportation systems, operations research, and robotics. 
\end{IEEEbiography}
\vspace{-3em}
\begin{IEEEbiography}
[{\includegraphics[width=1in,height=1.25in, clip,keepaspectratio]{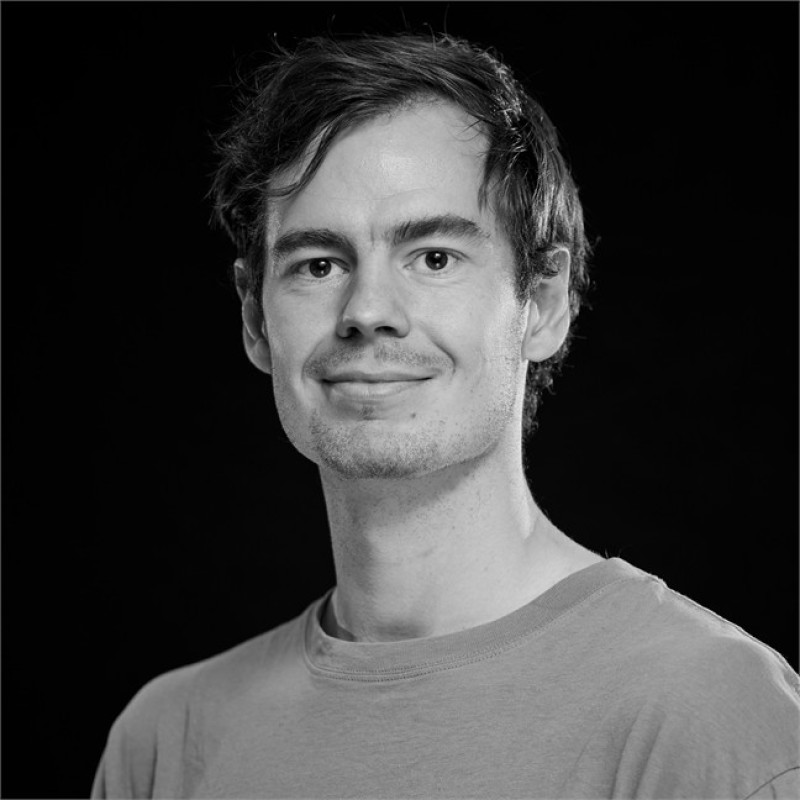}}]{Mathias Niemann Tygesen} holds a PhD from the Technical University of Denmark, where he specialized in Spatio-Temporal Machine Learning for public transport. Following a PostDoc on Reinforcement Learning for dynamic road pricing, he now works on demand forecasting in the retail fashion industry, applying advanced machine learning to optimize supply chains and inventory management. His work focuses on translating cutting-edge research into practical solutions across transportation and retail sectors.
\end{IEEEbiography}
\vspace{-3em}
\begin{IEEEbiography}
[{\includegraphics[width=1in,height=1.25in, clip,keepaspectratio]{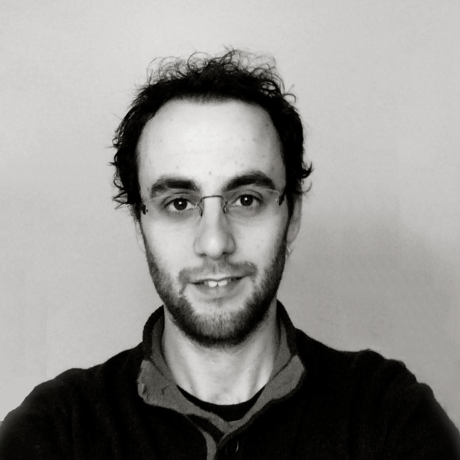}}]{Filipe Rodrigues} is associate professor at the Technical University of Denmark (DTU) in the Machine Learning for Smart Mobility (MLSM) lab, where his research is primarily focused on machine learning models for understanding and optimizing urban mobility and human behavior. Previously, he was a H.C. Ørsted / Marie-Skłodowska Curie Actions (COFUND) postdoctoral fellow, also at DTU, working on spatio-temporal models of mobility demand with emphasis on modeling uncertainty and the effect of special events. He has published more than 60 articles in leading conferences and journals in both transportation and machine learning. His research interests span machine learning, reinforcement learning, intelligent transportation systems, and urban mobility.
\end{IEEEbiography}







\end{document}